\documentclass[11pt]{article} 
\usepackage{epsfig}
\usepackage{subfigure}
\usepackage{graphics}
\usepackage{rotating}
\usepackage{amssymb}

\def\tr{\top}
\def\ov{\overline}
\def\til{\widetilde}
\def\lla{\left\langle}
\def\rra{\right\rangle}

\newcommand{\Bmath}[1]{\mbox{\bf {#1}}}

\def\x{{\Bmath{x}}}

\def\w{{\Bmath{w}}}
\def\m{{\Bmath{m}}}
\def\a{{\Bmath{a}}}

\def\s{{\Bmath{s}}}

\def\w{{\Bmath{w}}}
\def\u{{\Bmath{u}}}

\def\v{{\Bmath{v}}}
\def\e{{\Bmath{e}}}

\def\M{{\Bmath{M}}}
\def\P{{\Bmath{P}}}
\def\R{{\Bmath{R}}}

\def\B{{\Bmath{B}}}

\def\A{{\Bmath{A}}}

\def\C{{\Bmath{C}}}

\def\T{{\Bmath{T}}}
\def\W{{\Bmath{W}}}
\def\Q{{\Bmath{Q}}}

\def\V{{\Bmath{V}}}
\def\I{{\Bmath{I}}}
\def\R{{\Bmath{R}}}
\def\S{{\Bmath{S}}}

\def\cX{{\cal X}}

\newcommand{\RR}{{\mathbb{R}}}
\newcommand{\CC}{{\mathbb{C}}}

\newcommand{\NN}{{\mathbb{N}}}

\newtheorem{definition}{Definition}[section]
\newtheorem{theorem}[definition]{Theorem}

\newtheorem{corollary}[definition]{Corollary}
\newcommand{\proof}{{\underline{Proof:} \quad}}
\newcommand{\ep}{\mbox{}\hfill $\Box$}

\begin{document}

\title{Dynamical Systems as Temporal Feature Spaces}

\author{\bf  Peter Tino\\
	\it School of Computer Science\\
        \it The University of Birmingham \\
	\it Birmingham B15 2TT, UK\\
	{\tt P.Tino@cs.bham.ac.uk} 
      }

\maketitle

\begin{abstract}
Parameterized state space models in the form of recurrent networks are often used in machine learning to learn from data streams exhibiting temporal dependencies. To break the black box nature of such models it is important to understand the dynamical features of  the input driving time series that are formed in the state space. 
We propose a  framework for rigorous analysis of such state representations in vanishing memory state space models such as echo state networks (ESN). In particular, we consider the state space a temporal feature space and the readout mapping from the state space a kernel machine operating in that feature space. We show that: 
{\bf (1)}
The usual ESN strategy of randomly generating input-to-state, as well as state coupling leads to shallow memory time series representations, corresponding to cross-correlation operator with fast exponentially decaying coefficients;
{\bf (2)}
Imposing symmetry on dynamic coupling yields a constrained dynamic kernel matching the input time series with straightforward exponentially decaying motifs or exponentially decaying motifs of the highest frequency;  
{\bf (3)}
Simple cycle high-dimensional reservoir topology specified only through two free parameters can implement deep memory dynamic kernels with a rich variety of matching motifs.
We quantify richness of feature representations imposed by dynamic kernels and demonstrate that for dynamic kernel associated with cycle reservoir topology, 
the kernel richness undergoes a phase transition close to the edge of stability.
\end{abstract}

\section{Introduction}
When dealing with time series data, techniques of machine learning and signal processing must account in some way for temporal dependencies in the data stream. One popular option is to impose a parametric state-space model structure in which the state vector is supposed to dynamically code for the  input time series processed so far and the output is determined through a static readout from the state. Recurrent neural networks (e.g. \cite{Downey:2017:PSR:3295222.3295354}), 
Kalman filters \cite{Kalman1960} or hidden Markov models \cite{baum1966} represent just {a} few examples of this approach. 
In some cases the state space and transition structure is (at least partially) imposed based on the relevant prior knowledge \cite{Yoon09}, but usually it is learnt from the data along with the readout map. In {the}  case of uncountable state space and non-linear state dynamics, the use of gradient methods in learning the state transition dynamics is hampered by the well known ``information latching problem" \cite{Ben93a}.
As temporal sequences increase in length, the influence of early
components of the sequence have less impact on the network output.
This causes the partial gradients, used to update the weights, to
(exponentially) shrink
to zero as the sequence length increases.
Several approaches have been suggested to overcome this challenge, e.g.
\cite{Bengio94,Hochreiter97,Li2019}. 

One possibility to avoid having to train the state transition part in a state space model is to simply initialize it randomly to a `sensible' fading memory dynamic filter and only train the static readout part of the model. Models following this philosophy \cite{Jaeger2001,Maass2002,Tino2001} have been termed ``reservoir computation (RC) models''
\cite{Lukoservicius2009}. 
Perhaps the simplest form of a RC model is the Echo State Network (ESN) \cite{Jaeger2001,Jaeger2002,jaeger2002a,Jaeger2004}.
Briefly, ESN is a recurrent neural network with a non-trainable state transition part (reservoir) and a simple trainable linear readout. Connection weights in the ESN reservoir, as well as the input weights are randomly generated. 
The reservoir weights are scaled
so as to ensure the {\em ``Echo State Property''} (ESP): the reservoir state is an {\em ``echo''} of the entire input history {and does not depend on the initial state}. {Scaling reservoir weights so that the largest singular value is smaller than 1 makes the reservoir dynamics contractive and guarantees the ESP. In practice, sometimes it is the spectral radius that guides the scaling. In this case, however, spectral radius $<1$ does not guarantee the ESP.}

ESNs has been successfully applied in a variety of 
tasks \cite{Jaeger2004,Skowronski2006,Bush2005,Tong2007}. 
Many extensions of the classical ESN have been suggested in the literature, e.g. deep ESN \cite{GALLICCHIO201787}, intrinsic
plasticity \cite{Schrauwen2008,Steil2007}, decoupled reservoirs
\cite{Xue2007}, 
leaky-integrator reservoir units \cite{Jaeger2007},  filter neurons with delay-and-sum readout \cite{holzmann2009} etc.
However, there are properties of the reservoir that are poorly understood \cite{Xue2007} and 
specification of the reservoir and input connections require numerous trails and even luck \cite{Xue2007}.
Furthermore, imposing a constraint on spectral radius or largest singular value of the reservoir matrix is a weak tool to properly set the reservoir parameters \cite{Ozturk2007}.  Finally, random
connectivity and weight structure of the reservoir is unlikely to be optimal and such a setting prevents us from providing a clear and systematic insight into the reservoir dynamics organization \cite{Ozturk2007,Rodan10}. 
\cite{Rodan10} demonstrated that even an extremely simple setting of a high-dimensional state space structure governed by only two free parameters set deterministically can yield modelling capabilities on par with other ESN architectures. However, a deeper understanding of why this is so has been missing.

In order to theoretically understand the workings of parameterized state space models as machine learning tools to process and learn from temporal data, there has been a lively research activity to formulate and assess different aspects of computational power and information processing capacity in such systems (e.g. \cite{Dambre2012,Obst2014,Hammer01,Hammer03,siegelmann94analog,Tino03NC}). For example, tools of information theory have been used to assess information storage or transfer within systems of this kind \cite{Lizier2007,Lizier2012,Obst2010,Bossomaier2016}. 
To specifically characterize capability of input-driven dynamical systems to keep in their state-space information about past inputs,
several memory quantifiers were proposed, for example ``short term memory capacity" \cite{Jaeger2002} and ``Fisher memory curve" \cite{Sompolinsky2008,Tino2018_Fisher}. Even though those two measures have been developed from completely different perspectives, deep connections exist between them \cite{Tino2013}. The concept of memory capacity, originally developed for univariate input streams, was generalized to multivariate inputs in \cite{Grigoryeva2016}. Couillet et al. \cite{Couillet2016} rigorously studied  mean-square error of linear dynamical systems used as dynamical filters in regression tasks and suggested memory quantities that generalize the short term memory capacity and Fisher memory curve measures.
Finally,  \cite{Ganguli2010} showed an interesting connection between memory in dynamical systems and their capacity to perform dynamical compressed sensing of past inputs.

In this contribution, we suggest a novel framework for characterizing richness of dynamic representations of input time series in the form of states of a dynamical system, which is the core part of any state space model used as a learning machine. 
{Our framework is based on the observation that the idea of fixed dynamic reservoir with simple static linear mapping build on top of it strikingly resembles the philosophy of kernel machines \cite{Legenstein2007}. There, the inputs are transformed using a fixed mapping (usually only implicitly defined) into a feature space that is "rich enough" so that in that space it is sufficient to train linear models only. The key tool for building linear models in the feature space is the inner product. One can grasp workings of a kernel machine only by understanding of how the data is mapped to the feature space and what "data similarity" in the original space means when expressed as the inner product in the feature space. We will view the reservoir state space as a "temporal feature space" in which the linear readout is operating. In this view, the input time series seen by the reservoir model results in a state that codes all history of the presented input items so far and thus forms a feature representation of the time series. Different forms of coupling in the reservoir dynamical system will result in different temporal feature spaces with different feature representations of input time series, implying different notions of similarity between time series, expressed as inner products of their feature space representations. We will ask if and how the feature spaces differ in cases of traditional randomly generated reservoir models, as well as more constrained reservoir constructions studied in the literature.
}

{Since RC models are input-driven driven non-autonomous dynamical systems, theoretical studies linking their information processing capabilities to the reservoir  coupling structures have been performed mostly in the context of linear dynamics, e.g.  \cite{Sompolinsky2008,Couillet2016-SSP,Couillet2016,Tino2018_Fisher}.
While such studies are of interest by themselves, in the context of the present work, studying linear dynamics can shed light on a wide class of RC models whose approximation capabilities equal those of non-linear systems.
In particular, 
Grigoryeva, Gonon and Ortega recently proved a series of important results concerning  universality of RC models  
\cite{Grigoryeva2018,GRIGORYEVA2018_NN,Gonon19}. The universality can be obtained even if the state transition dynamics is linear, provided the readout map is polynomial (or a neural network)\footnote{
Universal approximation capability was first established in the $L^\infty$
sense for deterministic, as well as almost surely uniformly
bounded stochastic inputs \cite{Grigoryeva2018}. This was later extended in \cite{Gonon19} to $L^p$, $1\le p<\infty$ and not necessarily almost surely uniformly
bounded stochastic inputs.}.
However, universality is a property of a whole {\em family} of RC models. For appropriate classes of filters\footnote{
transforming semi-infinite input sequences into outputs} 
and input sources, it guarantees that for any filter and approximation precision, there exists a RC model approximating the filter
to that precision. This is an existential statement that does relate individual filters to their approximant RC models. Our new framework will enable us to reason about what kind of RC model setup is necessary if filters with deeper memory were to be approximated.
In particular, we will first investigate properties of linear dynamical readout kernels obtained on top of linear dynamical systems. Crucially, memory properties of such kernels can not be enhanced by moving from linear to polynomial static readout kernels. 
Loosely speaking, if feature representation $\x$ of a time series $\u$ captures properties of $\u$ only up to some look-back time $t-\tau_0$ from the last observation time $t$, then no nonlinear transformation $\gamma$ of $\x$ can prolong memory $\tau_0$ in the feature representation $\gamma(\x)$ of $\u$.
Hence, we will be able to make statements regarding appropriate settings of the linear dynamics that are necessary for universal approximation of deeper memory filters. 
}

The paper has the following organization: In section \ref{sec:prelim} we set the scene and outline the main intuitions driving the work. Section \ref{sec:temp_kernel} formally introduces the notion of temporal kernel and provides some useful properties of the kernel to be used further in our study.
In section \ref{sec:motifs} we will setup basic tools for characterizing dynamic kernels - motifs and their corresponding motif weights. Starting from section \ref{sec:rand_W}, we will analyze dynamic kernels corresponding to different settings of the dynamical system. In particular, 
dynamical kernels associated with fully random, symmetric and highly constrained coupling of the dynamical system are analyzed in sections  \ref{sec:rand_W},  \ref{sec:rand_sym_W} and
\ref{sec:scr_W}, respectively. We provide examples illustrating the developed theory and compare the motif richness of different parameter settings of the dynamical system in section 
\ref{sec:examples}. The paper finishes with discussion and conclusions in section \ref{sec:Conclusion}.

\section{Preliminary concepts and intuitions}
\label{sec:prelim}

We consider fading memory state space models with linear input driven dynamics in an $N$-dimensional state space and univariate inputs and outputs. 
Note that in the ESN metaphor, the state dimensions correspond to reservoir units coupled to the input $u(t)$ via $N$-dimensional weight vectors $\w \in \RR^N$. Denoting the state vector at time $t$ by $\textbf{x}(t) \in \RR^N$, the dynamical system evolves as
\begin{equation}
\x(t) = \W \ \x(t-1) + \w \ u(t),
\label{eq:state}
\end{equation}
where $\W \in \RR^{N \times N}$ is a $N \times N$ weight matrix providing the dynamical coupling.
In state space models, 
the output $y(t)$ is often determined solely based on the current state $\x(t)$ through a readout function $h$:
\begin{equation}
y(t) = h(\x(t)).
\label{eq:readout}
\end{equation}
The readout map $h$ is typically trained (offline or online) by minimizing
the (normalized) mean square error between the targets and reservoir readouts $y(t)$. 

{Denote the set of natural numbers (including zero) by $\NN_0$.}
In this contribution, we study how the dynamical system (\ref{eq:state})
extracts potentially useful information about the {left infinite input time series $..., u(t-2), u(t-1), u(t)$,
$u(-j) \in \RR, j \in \NN_0$,}
 in its state $\x(t) \in \RR^N$, since it is only the state $\x(t)$ that will be used to produce the predictive output $y(t)$ upon seeing the input time series up to time $t$. In particular, we will consider readout maps constructed in the framework of kernel machines.
For example, in the case of linear Support Vector Machine (SVM) regression, the readout from
the state space at time $t$ has the form 
\begin{equation}
y(t) = h(\x(t)) = \sum_{i} \beta_i\ \lla\x(t_i),\x(t)\rra + b,
\label{eq:svm_temp_kernel}
\end{equation}
where $\beta_i\in \RR$ and $b \in \RR$ are weight coefficients and bias term, respectively and $\x(t_i)$ are support state vectors observed at important ``support time instances" $t_i$.
In the spirit of state space modelling discussed above, we consider the state $\x(t') \in \RR^N$ reached after observing the time series $..., u(t'-2), u(t'-1), u(t')$, the feature state space representation of that time series.
Hence (\ref{eq:svm_temp_kernel}) can also be written as
\begin{equation}
y(t) = \sum_{i} \beta_i\ K([...u(t_i-1), u(t_i)], [...u(t-1), u(t)]) + b,
\label{eq:svm_temp_kernel1}
\end{equation}
where $K(\cdot,\cdot)$ is a time series kernel 
associated with the dynamical system (\ref{eq:state}),
\begin{equation}
K([...u(t_i-1), u(t_i)], [...u(t_j-1), u(t_j)]) = 
\lla\x(t_i),\x(t_j)\rra.
\label{eq:dyn_kernel}
\end{equation}
In this context, the support time instances $t_i$ can be viewed as end times of the ``support time series" $..., u(t_i-2), u(t_i-1), u(t_i)$ observed in the past and deemed ``important" for producing the outputs by the training algorithm trained on the history of the time series before the time step $t$.

\begin{figure}
\centering
\includegraphics[width=14cm]{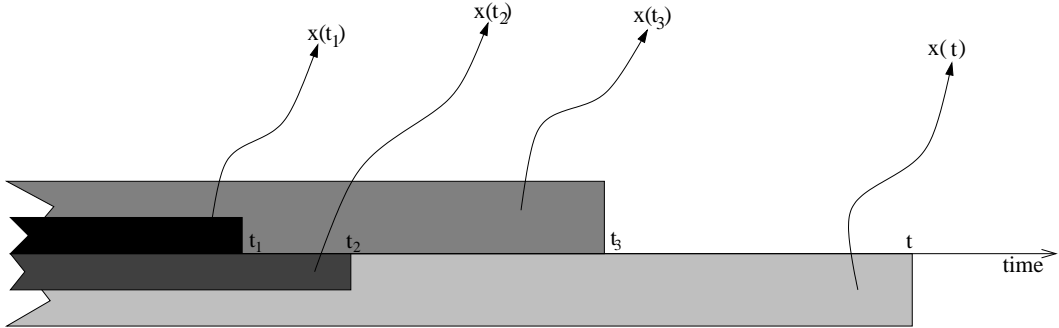}
\caption{Illustration of the workings of kernel machine producing an output
at time $t$ after observing $..., u(t-1), u(t)$. The time series 
$..., u(t-1), u(t)$ is compared with the three support time series 
$(..., u(t_i-1), u(t_i))$, $i=1,2,3$, by evaluating dot products between their
feature space representations $\x(t)$ and $\x(t_i)$.
}
\label{fig:illustration_temp_kernel}
\end{figure}

The suggested viewpoint is illustrated in figure \ref{fig:illustration_temp_kernel}. There are three support time series $(..., u(t_1-2), u(t_1-1), u(t_1))$, $(..., u(t_2-2), u(t_2-1), u(t_2))$ and 
$(..., u(t_3-2), u(t_3-1), u(t_3))$ represented through the states $\x(t_1)$, $\x(t_2)$ and $\x(t_3)$, respectively. To evaluate the output at time $t$, the current feature space representation $\x(t)$ of $...,u(t-2),  u(t-1), u(t)$ is compared with feature space representations $\x(t_i)$, $i=1,2,3$, of the support time series through dot products. 

We will next {formalize these intuitions} and then investigate the properties of state space feature representations of time series by dynamical systems. In particular, we will be interested in how different forms of dynamic coupling $\W$ influence richness of such feature representations and how they map to properties of the corresponding temporal kernel.

\section{Temporal kernel defined by dynamical system}
\label{sec:temp_kernel}
Without loss of generality, we will study feature state space representations under the dynamical system (\ref{eq:state}) of left-infinite time series $..., u(-2), u(-1), u(0), \ \ \ u(-j) \in \RR, \ j \in \NN_0$. 
{We will assume that the largest singular value $\nu$
of the dynamic coupling $\W$ is strictly less than 1}, making the dynamics (\ref{eq:state}) contractive. This means that the echo state property is fulfilled and for sufficiently long past horizons $\tau \gg 1$, the influence of initial state $\x(-\tau)$ on the feature representation of 
\[
u(-\tau+1), u(-\tau+2),...,u(-1),u(0)
\]
is negligible.
{
Note that $\nu<1$ is a sufficient condition for the echo state property, but the property may actually be achieved  under milder conditions, especially when particular input streams are considered (for formal treatment and further details see e.g. \cite{YILDIZ20121,Manjunath:2013}).  In this contribution we use $\nu<1$, since it allows us {\bf (1)} to consider arbitrary input streams over a bounded domain (the ESP is always guaranteed) and {\bf (2)} to explicitly bound, in terms of properties of $\W$, the norm of dynamical states, as well as the extent to which the initial state  
influences the temporal kernel.}

More formally, given a past horizon $\tau \gg 1$, we will represent the time series $u(-\tau+1), u(-\tau+2),...,u(-1),u(0)$ 
as a vector $\u(\tau) = (u_1, u_2, ..., u_\tau)^\tr \in \RR^\tau$,
where $u_i = u(-i+1), i=1,2,...\tau$. In other words
$\u(\tau) = (u(0), u(-1), ..., u(-\tau+1))^\tr$.

Consider a state \ $\x(-\tau) \in \RR^N$ at time $-\tau$. After seeing
the input series $u(-\tau+1), u(-\tau+2),...,u(-1),u(0)$, the new state
of the dynamics (\ref{eq:state}) will be\footnote{
$\W^0=\I_{N \times N}$, the $N \times N$ identity matrix.}
\[
\x(0) = \W^\tau \x(-\tau) + \sum_{j=1}^{\tau} u(j-\tau) \W^{\tau-j} \w.
\]
As discussed in the previous section, the state $\x(0)$ reached from the initial condition $\x(-\tau)$ after seeing $\u(\tau)$ codes for information content in $\u(\tau)$ and will be considered the ``feature space representation" of $\u(\tau)$ through the dynamical system (\ref{eq:state}):
\begin{eqnarray}
\phi(\u(\tau); \ \x(-\tau)) 
&=& \x(0)
\nonumber \\
&=& \W^\tau \x(-\tau) + \sum_{i=1}^{\tau} u(1-i) \W^{i-1} \w
\nonumber \\
&=& \W^\tau \x(-\tau) + \sum_{i=1}^{\tau} u_i \W^{i-1} \w.
\label{eq:phi_general}
\end{eqnarray}

Given two time series at past horizon $\tau$ represented through
$\u(\tau) = (u_1, u_2, ..., u_\tau)^\tr$ and $\v(\tau) = (v_1, v_2, ..., v_\tau)^\tr$, the temporal kernel defined by dynamical system (\ref{eq:state})
evaluated on $\u(\tau)$ and $\v(\tau)$ reads:
\begin{equation}
K(\u(\tau),\v(\tau);\ \x(-\tau)) = \lla \phi(\u(\tau);\ \x(-\tau)),\phi(\v(\tau); \ \x(-\tau)) \rra.
\label{eq:K_old}
\end{equation}

We will now show that, as expected given the contractive nature of (\ref{eq:state}), 
{for sufficiently long past time horizons $\tau \gg 1$ on input streams over bounded domain}\footnote{
{It is common in the ESN literature to consider input streams over a bounded domain (e.g. \cite{Jaeger2001}). 
In the recent work on universality of ESNs \cite{Grigoryeva2018} consider 
almost surely uniformly bounded stochastic inputs.
This is further relaxed in \cite{Gonon19}.}},
 the kernel evaluation is insensitive to the initial condition $\x(-\tau)$. This will allow us to simplify the presentation by setting $\x(-\tau)$ to the origin in the rest of the paper.

\vskip 0.5cm
\begin{theorem}
\label{Thm:K_0}
{
Consider the dynamical system (\ref{eq:state}) driven by time series over a bounded domain $[-U,U]$, $0<U<\infty$, with a past time horizon $\tau>1$. {Assume that the largest singular value $\nu$ of the dynamic coupling $\W$} is strictly smaller than 1 and that the norm of the input coupling $\w$ satisfies $\|\w\| \le B$.  Assume further that the norm of the initial condition is upper bounded by $\| \x(-\tau)\| \le A(\tau) = c \cdot \zeta^{-\tau}$, where $\nu < \zeta <1$
and
 $c>0$ is a large enough positive constant satisfying
\begin{equation}
c \ge \frac{B \cdot U}
{(1-\nu) \cdot \left(1- \frac{\nu}{\zeta} \right)}.
\label{eq:c}
\end{equation}
Then, for any
$\u(\tau),\v(\tau) \in [-U,U]^\tau$, it holds
\[
K(\u(\tau),\v(\tau);\ \x(-\tau)) = K(\u(\tau),\v(\tau);\ {\mathbf 0}) + \epsilon,
\]
where
\begin{equation}
- \eta^\tau \left[\frac{2 c}{1-\nu} \cdot B \cdot U\right] 
\le \ \epsilon \
\le
\eta^\tau 
\left[
c^2 \ \eta^\tau  + \frac{2 c}{1-\nu}  \cdot B \cdot U
\right],
\label{eq:bound_eps}
\end{equation}
with $\eta = \nu/ \zeta <1$.
}
\end{theorem}

\vskip 0.5cm
\proof
Note that
\[
\phi(\u(\tau);\ \x(-\tau)) = \W^\tau \x(-\tau) + \phi(\u(\tau);\ {\mathbf 0})
\]
and therefore, denoting 
\begin{equation}
\phi(\u(\tau);\ {\mathbf 0}) = \sum_{i=1}^{\tau} u_i \W^{i-1} \w
\label{eq:phi_0}
\end{equation}
 by 
$\phi_0(\u(\tau))$, 
we have
\begin{eqnarray}
K(\u(\tau),\v(\tau);\ \x(-\tau))
&=& 
\lla \W^\tau \x(-\tau) + \phi_0(\u(\tau)),
\W^\tau \x(-\tau) + \phi_0(\v(\tau)) \rra
\nonumber \\
&=& \| \W^\tau \x(-\tau) \|^2_2
+ \lla \W^\tau \x(-\tau), \phi_0(\u(\tau)) + \phi_0(\v(\tau)) \rra
\nonumber \\
&+& K(\u(\tau),\v(\tau);\ {\mathbf 0}),
\nonumber
\end{eqnarray}
where
\[
K(\u(\tau),\v(\tau);\ {\mathbf 0}) = \lla \phi_0(\u(\tau)),\phi_0(\v(\tau)) \rra 
\]
is the dynamic kernel evaluated using initial condition $\x(-\tau)$ set to the origin ${\mathbf 0}$. We have,
\begin{equation}
\| \W^\tau \x(-\tau) \|^2_2 \le \nu^{2 \tau} \cdot {(A(\tau)})^2.
\label{eq:p1}
\end{equation}
Note that
\[
\lla \W^\tau \x(-\tau), \phi_0(\u(\tau))  \rra
\le
\|\W^\tau {\x(-\tau)}\| \cdot \|\phi_0(\u(\tau))\|
\]
and (see (\ref{eq:phi_0}))
\begin{eqnarray}
\|\phi_0(\u(\tau))\| 
&\le&
\sum_{i=1}^{\tau} \nu^{i-1} \cdot U \cdot \|\w\|
\nonumber \\
&\le&
B \cdot U \cdot  \frac{1}{1-\nu},
\label{eq:p2}
\end{eqnarray}
yielding
\[
\lla \W^\tau \x(-\tau), \phi_0(\u(\tau))  \rra
\le
\frac{\nu^\tau}{1-\nu} {\cdot A(\tau)} \cdot B \cdot U.
\]
We thus have
\[
K(\u(\tau),\v(\tau);\ \x(-\tau)) = K(\u(\tau),\v(\tau);\ {\mathbf 0}) + \epsilon,
\]
with
\[
\epsilon 
\le 
\nu^\tau 
\left[
\nu^\tau ({A(\tau))}^2 + \frac{2}{1-\nu}  {\cdot A(\tau)} \cdot B \cdot U
\right].
\]

To evaluate the lower bound on $\epsilon$, note that
\begin{eqnarray}
\lla \W^\tau \x(-\tau), \phi_0(\u(\tau))  \rra
&\ge&
- \|\W^\tau {\x(-\tau)}\| \cdot \|\phi_0(\u(\tau))\|
\nonumber \\
&\ge&
\frac{-\nu^\tau}{1-\nu} {\cdot A(\tau)} \cdot B \cdot U.
\nonumber
\end{eqnarray}
Since, trivially, $\| \W^\tau \x(-\tau) \|^2_2 \ge 0$, we have
\[
\epsilon 
\ge 
- \frac{2\nu^\tau}{1-\nu} {\cdot A} \cdot B \cdot U.
\]

We have thus obtained,
\[
- \nu^\tau \left[\frac{2}{1-\nu} {\cdot A(\tau)} \cdot B \cdot U\right] 
\le \epsilon
\le
\nu^\tau 
\left[
\nu^\tau ({A(\tau)})^2 + \frac{2}{1-\nu} {\cdot A(\tau)} \cdot B \cdot U
\right],
\]
{which is equivalent to (\ref{eq:bound_eps}).

In order to reconcile this setting with the dynamics (\ref{eq:state}),
consider a past horizon $\tau+\tau_0$ for some additional look-back time $\tau_0 \ge 1$. We require,
\[
\| \x(-\tau) \| \le A(\tau) = c \cdot \zeta^{-\tau}
\ \ \ \hbox{and} \ \ \ \| \x(-\tau-\tau_0) \| \le A(\tau+\tau_0) = c \cdot \zeta^{-\tau -\tau_0}.
\]
But from the dynamics (\ref{eq:state}), we also have (see eqs.(\ref{eq:phi_general}), (\ref{eq:p1}) and (\ref{eq:p2})),
\begin{eqnarray}
\| \x(-\tau) \| 
& \le &
\| \W^{\tau_0} \ \x(-\tau - \tau_0)  \| 
+ \frac{B \cdot U}{1-\nu}
\nonumber \\
& \le &
\nu^{\tau_0} \cdot A(\tau+\tau_0) + \frac{B \cdot U}{1-\nu}.
\end{eqnarray}
We would like the norm of the state $\x(-\tau)$ reached from the initial
state $\x(-\tau-\tau_0)$ (bounded in norm by $A(\tau+\tau_0)$) to be within the required bound $A(\tau)$. In other words, we would like
\begin{equation}
A(\tau+\tau_0) \cdot \nu^{\tau_0} + \frac{B \cdot U}{1-\nu}
< c \cdot \zeta^{-\tau}
\label{eq:ineq1}
\end{equation}
to hold. Using $A(\tau+\tau_0) = c \cdot \zeta^{-\tau -\tau_0}$,
we conclude that the inequality (\ref{eq:ineq1}) holds when
\begin{equation}
c > \zeta^\tau \cdot \frac{B \cdot U}{(1-\nu)}
 \cdot \frac{1}{1 - \left(\frac{\nu}{\zeta}\right)^{\tau_0}}.
\label{eq:const_c}
\end{equation}
Since $0<\eta = \nu/\zeta <1$, for $\tau, \tau_0 \ge 1$,
\[
\zeta^\tau \cdot \frac{B \cdot U}{(1-\nu)}
 \cdot \frac{1}{1 - \eta^{\tau_0}}
\ < \
\frac{B \cdot U}{(1-\nu) \cdot (1 - \eta)}
\]
we have that 
the inequality (\ref{eq:const_c}) is definitely satisfied when
\[
c > \frac{B \cdot U}{(1-\nu) \cdot (1 - \frac{\nu}{\zeta})}.
\]
}
\ep

Theorem \ref{Thm:K_0} formally states that because the dynamical system (\ref{eq:state}) is contractive, the influence of the initial condition $\x(-\tau)$ on the kernel value $K(\u(\tau),\v(\tau);\ \x(-\tau))$
decays exponentially with the past time horizon $\tau$. For sufficiently long past time horizons $\tau \gg 1$ we can thus set 
$\x(-\tau) = {\mathbf 0}$. Hence, in the rest of this study we will assume $\tau \ge N$ and (unless necessary) we will drop specific reference to $\tau$ by writing $\u$ instead of $\u(\tau) \in \RR^\tau$. In fact, it will be easier to think of time horizons in units of $N$, so that $\tau = \ell\cdot N$, for some sufficiently large integer $\ell >1$.
Furthermore, we will
refer to $\phi(\u(\tau);\ {\mathbf 0})$ and $K(\u(\tau),\v(\tau);\ {\mathbf 0})$ simply as $\phi(\u)$ and $K(\u,\v)$, respectively.

\section{Temporal kernel and its motifs}
\label{sec:motifs}
In the previous section we established that the temporal kernel associated with dynamical system (\ref{eq:state}) and acting on time series with past time horizon $\tau \gg 1$ is defined as
\begin{equation}
K(\u,\v) = \lla \phi(\u),\phi(\v) \rra.
\label{eq:K}
\end{equation}
{In order to analyze the action of $K(\u,\v) $ on time series $\u, \v$, we need to find its expression directly in terms of $\u$ and $\v$. The next theorem shows that there exists a matrix $\Q$ of rank at most $N$ that acts as a metric tensor on a subspace of $\RR^\tau$ (of dimensionality at most $N$), so that $K(\u,\v) $ can be expressed as  a quadratic form $\u^\tr  \Q \ \v$. This will allow us to study properties of  $K(\u,\v)$ by analyzing the associated metric tensor $\Q$.
}

\vskip 0.5cm
\begin{theorem}
\label{Thm:Q}
Consider the dynamical system (\ref{eq:state}) of state dimensionality $N$ and a dynamic coupling $\W$ with {largest singular value $0<\nu<1$}. Let $K(\u,\v)$ (\ref{eq:K}) be the temporal kernel associated with system (\ref{eq:state}). Then for any
$\u,\v \in \RR^\tau$,
\[
K(\u,\v) = 
\u^\tr  \Q \ \v =
\lla \u,\v \rra_{\Q},
\]
where $\Q$ is a symmetric, positive semi-definite $\tau \times \tau$ matrix
of rank  $N_m=\hbox{\rm rank}(\Q) \le N$ and elements
\begin{equation}
Q_{i,j} = \w^\tr \left(\W^\tr\right)^{i-1} \ \W^{j-1} \ \w, \ \ \ i,j=1,2,...,\tau.
\label{eq:Q_ij}
\end{equation}
The upper bound on absolute values of $Q_{i,j}$ decays exponentially with increasing time indices $i,j=1,2,...,\tau$, as
\begin{equation}
|Q_{i,j}| \ {\le} \ \nu^{i+j-2} \ \| \w \|_2^2.
\label{eq:abs_Q_ij}
\end{equation}

\end{theorem}

\vskip 0.5cm
\proof
First, we write
\begin{eqnarray}
K(\u,\v) 
&=& 
\lla \phi(\u),\phi(\v) \rra
\nonumber \\
&=& 
\lla 
\sum_{i=1}^{\tau} u_i \W^{i-1} \w,
\sum_{j=1}^{\tau} v_j \W^{j-1} \w
\rra 
\ \ \ \ \hbox{(eq. (\ref{eq:phi_0})})
\nonumber \\
&=& 
\sum_{i,j=1}^\tau u_i \ v_j \ \lla 
\W^{i-1} \w,
\W^{j-1} \w
\rra 
\nonumber \\
&=& 
\sum_{i,j=1}^\tau u_i \ v_j \ Q_{i,j}
\nonumber \\
&=& 
\u^\tr  \Q \ \v.
\nonumber
\end{eqnarray}

Second, $\phi(\u)$ can be written as
$\phi(\u) = \Phi  \u$,
where $\Phi$ is an $N \times \tau$ matrix whose $i$-th column is equal to 
$\W^{i-1} \w$. Hence,
$K(\u,\v) = \u^\tr \Phi^\tr \Phi \ \v$ and $\Q = \Phi^\tr \Phi$ is symmetric positive semi-definite with rank at most $N \le \tau$.

Finally,
since $\| \W^i \w\| \le \|\W^i\| \|\w\| \le \nu^i \|\w\|$, we have
\begin{eqnarray}
|Q_{i,j}|  
&=& 
|\lla \W^{i-1} \w, \W^{j-1} \w \rra|
\nonumber \\
&\le& 
\|\W^{i-1} \w\|_2 \cdot \|\W^{j-1} \w\|_2 
\nonumber \\
&\le& 
\nu^{i+j-2} \ \| \w \|_2^2.
\nonumber
\end{eqnarray}

\ep
\vskip 0.5cm

Note that $K(\cdot,\cdot)$ is a semi-inner product on $\RR^\tau$. In other words, time series $\u \in \ker{(\Q})$ from the kernel of the linear operator $\Q$ have zero length. It acts as an inner product in  the quotient of $\RR^\tau$ by $\ker{(\Q})$, 
 $\RR^\tau/\ker{(\Q})$ 
(image of $\Q$). Since this distinction is not crucial for our argumentation, in order not to unnecessarily complicate the presentation, (slightly abusing mathematical terminology) we will  refer to $K(\cdot,\cdot)$  as temporal kernel and to $\Q$ as the associated metric tensor.

Theorem \ref{Thm:Q} tells us that $K(\cdot,\cdot)$ is a fading memory temporal kernel and we can unveil its inner workings through eigen-analysis of $\Q$:
\begin{equation}
\Q = \M \Lambda \M^\tr,
\label{eq:Q_eigen}
\end{equation}
where the columns of $\M$ are the eigenvectors $\m_1, \m_2, ..., \m_\tau \in \RR^\tau$ 
of $\Q$ with the corresponding real non-negative eigenvalues $\lambda_1 \ge \lambda_2 \ge ... \ge \lambda_\tau$ arranged on the diagonal of the diagonal matrix $\Lambda$. 
Based on theorem \ref{Thm:Q}, there are $N_m \le N \le \tau$ eigenvectors 
$\m_i$ with positive eigenvalue $\lambda_i>0$. 

Given two time series $\u, \v \in \RR^\tau$ of past time horizon $\tau$, 
the temporal kernel value is
\begin{eqnarray}
K(\u,\v) 
&=& 
\u^\tr \ \Q \ \v
\nonumber \\
&=& 
\left(\Lambda^{\frac{1}{2}} \M^\tr \u\right)^\tr \ \Lambda^{\frac{1}{2}} \M^\tr \v.
\label{eq:K_m_t}
\end{eqnarray}
This has the following interpretation. In order to determine the kernel based ``similarity" $K(\u,\v)$ of two time time series $\u, \v \in \RR^\tau$, both  time series are first represented through a series of matching scores  with respect to a potentially small number of filters $\m_i$ ($N_m \le N \le \tau$) , weighted by $\lambda_i^{1/2}$:
\begin{equation}
\til \u = 
\left(\lambda_1^{1/2} \ \lla \m_1, \u \rra, 
 \lambda_2^{1/2} \ \lla \m_2, \u \rra, ...,
 \lambda_{N_m}^{1/2} \ \lla \m_{N_m}, \u \rra\right)^\tr \in \RR^{N_m}
\label{eq:tilde_u}
\end{equation}
and
\[
\til \v = 
\left(\lambda_1^{1/2} \ \lla \m_1, \v \rra, 
 \lambda_2^{1/2} \ \lla \m_2, \v \rra, ...,
 \lambda_{N_m}^{1/2} \ \lla \m_{N_m}, \v \rra\right)^\tr \in \RR^{N_m}.
\]
{Similarity between $\u \in \RR^\tau $ and $\v \in \RR^\tau$ is then evaluated as the degree to which both $\u$ and $\v$ match in the same way the highly weighted filters $\m_i$.} Hence, instead of direct matching of $\u$ and $\v$, as would be the case for $\lla \u, \v \rra$, we consider  $\u, \v$ ``similar" if $\lla \til \u, \til \v \rra$ is high, in other words, if {\em both} $\u$ and $\v$ match well a number of significant filters $\m_i$ of high weight $\lambda_i^{1/2}$.

{The matching scores $ \lla \m_i, \u \rra$ can be viewed as projections of the input time series $\u$ unto "prototypical" time series motifs $\m_i$ that characterize what features of the input time series are used by the kernel to assess their similarity. Loosely speaking, a temporal kernel employing a rich set of slowly decaying high-weight ("significant") motifs with deep memory will be able to perform more nuanced time series similarity evaluation than a kernel with a small set of highly constrained and fast decaying short memory motifs.  
In what follows we refer to $\m_1, \m_2, ..., \m_{N_m} \in \RR^\tau$ as {\em motifs of the temporal kernel} $K(\cdot,\cdot)$ with the associated {\em motif weights} given by $\lambda_1^{1/2} \ge \lambda_2^{1/2} \ge ... \ge \lambda_{N_m}^{1/2} >0$.}
In the light of the comments above, $K(\cdot,\cdot)$ acts as semi-inner product on $\RR^\tau$ and as inner product on the span of the motifs,
$\hbox{span}\{\m_1, \m_2, ..., \m_{N_m}\}$.

In the case of SVM regression, the readout output for a time series 
$\v \in \RR^\tau$, based on the state space representation of $\v$ through (\ref{eq:state}) would
have the form (see eq.(\ref{eq:svm_temp_kernel1}))
\[
\sum_{i} \beta_i\ K(\u_i,\v) + b,
\]
where  $\u_i\in \RR^\tau$ are the support vectors (``support time series").
This can be rewritten as a linear model $\a^\tr\v + b$
with weight vector $\a \in \RR^\tau$:
\begin{eqnarray}
\a^\tr
&=& 
\sum_{i} \beta_i\ \u_i^\tr \ \Q 
\nonumber \\
&=& 
\sum_i \beta_i \ \sum_{j=1}^M \lambda_j  
\ \lla \m_j,\u_i \rra
\ \m_j^\tr.
\end{eqnarray}
Free parameters of the output-producing function are the coefficients $\beta_i$ corresponding to the support time series 
$\u_i$. In contrast, motifs $\m_j$ and motif weights $\lambda_j^{1/2}$ are fixed by the dynamical system (\ref{eq:state}). Hence, whatever setting of the free parameters $\beta_i$ one can come up with, the inherent memory and time series structures one can access in past data in order to produce the output for a newly observed time series are determined by the richness and memory depth characteristics of the motif set $\{\m_j\}_{j=1}^{N_m}$.
In what follows we will take this viewpoint when analyzing temporal kernels corresponding to the dynamical system (\ref{eq:state}) for different types of state space coupling $\W \in \RR^{N \times N}$.

\section{Random dynamic coupling $\W$ with zero-mean i.i.d. entries}
\label{sec:rand_W}
{It has been common practice in the reservoir computation community to generate dynamic coupling $\W$ of ESNs randomly \cite{Lukoservicius2009}, typically with elements of $\W$ generated independently from a zero-mean symmetric distribution and then renormalized so that $\W$ has a desirable scaling property (e.g. certain spectral radius or largest singular value). In this section we investigate temporal kernels associated with such a ESN setting. We will see that the nature of motifs is remarkably stable (small set of shallow memory motifs), even though the couplings $\W$ are generated from a wide variety of distributions.

Consider a random matrix $\til \W$  with elements $\til W_{i,j}$, $i,j = 1,2,...,N$, generated i.i.d. from a zero-mean distribution with variance $\sigma_0^2>0$ and finite fourth moment. }Such a realization $\til \W \in \RR^{N \times N}$ will be {rescaled to the desired largest singular value} $\nu \in (0,1)$:
\[
\W = \frac{\nu}{\sigma_{max}(\til \W)} \til \W,
\]
where $\sigma_{max}(\til \W)$ is the maximum singular value of $\til \W$.

For large $N$, the largest eigenvalue of 
$N^{-1} \til \W^\tr \til \W$ converges to $4 \sigma_0^2$ almost surely \cite{Rudelson2010,Tino2018_Fisher}. Hence, the largest singular value of $N^{-1/2} \ \til \W$ approaches $2 \sigma_0$. It follows that for large $N$, $\sigma_{max}(\til \W)$ approaches $2 \sqrt{N} \sigma_0$.
Rescaling
\[
\W = \frac{\nu}{2 \sqrt{N} \sigma_0} \til \W
\]
can be equivalently thought of as generating $W_{i,j}$ i.i.d. from a zero-mean distribution with standard deviation
\begin{equation}
\sigma = \sigma_0 \frac{\nu}{2 \sqrt{N} \sigma_0} =
\frac{\nu}{2 \sqrt{N}}.
\label{eq:sigma}
\end{equation}

We would like to reason, under the assumption of high state space dimensionality $N$ of the dynamical system (\ref{eq:state}), about the properties of the metric tensor $\Q$ with elements $Q_{i,j}$ given by eq. (\ref{eq:Q_ij}).

To ease the mathematical notation, we denote the matrix $(\W^\tr)^i \ \W^j$ by 
$\A^{(i,j)}$. Hence,
\begin{equation}
Q_{i,j} = \w^\tr \ \A^{(i-1,j-1)} \ \w.
\label{eq:Q_A}
\end{equation}

\subsection{Diagonal elements of $\Q$} 
The first diagonal element of $\Q$, $Q_{1,1}$, is trivially equal to $\| \w \|^2_2$, 
so let us first concentrate on $\A^{(1,1)}$ corresponding to $Q_{2,2}$.

\begin{eqnarray}
\A^{(1,1)}_{j,j} 
&=& 
N \left[
\frac{1}{N} \sum_{i=1}^N W_{i,j}^2
\right]
\nonumber \\
&\approx& 
N \sigma^2
\nonumber \\
&=& 
\frac{\nu^2}{4}. \hskip 1cm \hbox{(see eq. (\ref{eq:sigma}))}
\label{eq:A^11_diag}
\end{eqnarray}

The off-diagonal terms of $\A^{(1,1)}$ get asymptotically small
as
\[
\A^{(1,1)}_{i,j} 
= 
N \left[
\frac{1}{N} \sum_{{k=1}}^N W_{k,i} \ W_{k,j} 
\right]
\approx 0
\]
{since for $i \neq j$, $W_{k,i}$ and $W_{k,j}$ are uncorrelated and generated from 
zero-mean distribution with standard deviation $\sigma = {\cal O}(1/\sqrt{N})$
(see (\ref{eq:sigma}))}. 
For large $N$ we can thus approximate $\A^{(1,1)}$ as
\begin{equation}
\A^{(1,1)} \approx \frac{\nu^2}{4} \ \I_{N \times N},
\label{eq:A^11}
\end{equation}
where $\I_{N \times N}$ is the identity matrix of rank $N$.

To approximate $\A^{(2,2)}$, we write
\begin{eqnarray}
\A^{(2,2)} 
&=& 
(\W^\tr)^2 \ \W^2
\nonumber \\
&=& 
\W^\tr \ \A^{(1,1)} \ \W
\nonumber \\
&\approx& 
\frac{\nu^2}{4} \ \W^\tr \ \W
\nonumber \\
&=&
\frac{\nu^2}{4} \A^{(1,1)}
\nonumber \\
&\approx& 
\left(\frac{\nu^2}{4}\right)^2 \ \I_{N \times N}.
\label{eq:A^22}
\end{eqnarray}

Proceeding inductively, we obtain
\begin{eqnarray}
\A^{(k,k)} 
&=& 
\W^\tr \ \A^{(k-1,k-1)} \ \W
\nonumber \\
&\approx& 
\frac{\nu^2}{4} \ 
\left(\frac{\nu^2}{4}\right)^{k-1} \ \I_{N \times N}
\nonumber \\
&=& 
\left(\frac{\nu^2}{4}\right)^k \ \I_{N \times N},
\ \ \ k=2,3,...,\tau-1.
\label{eq:A^kk}
\end{eqnarray}

We can thus approximate $Q_{j,j}$ as
\begin{eqnarray}
Q_{j,j}
&=& 
\w^\tr \ \A^{(j-1,j-1)} \ \w
\nonumber \\
&\approx& 
\left(\frac{\nu^2}{4}\right)^{j-1} \w^\tr \ \w
\nonumber \\
&=&
\left(\frac{\nu}{2}\right)^{2(j-1)} \|\w\|^2_2. 
\label{eq:approx_Q_jj}
\end{eqnarray}
Hence, the diagonal elements of $\Q$, necessarily non-negative since
$\A^{(j,j)}$ are positive semidefinite, decay much faster (exponentially so, by the factor of $4^{-(j-1)}$) than
the upper bound (\ref{eq:abs_Q_ij}) of theorem \ref{Thm:Q},
\begin{equation}
Q_{j,j} \approx
4^{-(j-1)} \ \nu^{2(j-1)} \ \|\w\|^2_2. 
\label{eq:B_ij_decay_rand_W}
\end{equation}

In particular, if all elements of the input coupling $\w$ have the same absolute value $w$ (with possibly different signs), we have
\begin{equation}
Q_{j,j} \approx 
N w^2 \left(\frac{\nu}{2}\right)^{2(j-1)}.
\label{eq:approx_Q_jj_w}
\end{equation}

\subsection{Off-diagonal elements of $\Q$} 
We now investigate terms $Q_{i,j}$ for $i \neq j$. Since Q is symmetric, without loss of generality we can assume $j>i$. Then,
\begin{eqnarray}
\A^{(i-1,j-1)}
&=& 
(\W^\tr)^{i-1} \ \W^{i-1} \ \W^{j-i}
\nonumber \\
&=&
\A^{(i-1,i-1)} \ \W^{j-i}
\nonumber \\
&\approx& 
\left(\frac{\nu}{2}\right)^{2(i-1)} \ \W^{j-i}
\label{eq:approx_A_ij}
\end{eqnarray}

The elements of $\A^{(i-1,j-1)}$ decay exponentially with increasing $i$ (deeper past in the time series). We will now approximate $\W^{j-i}$.

Concentrate first on the sub- and super-diagonal elements of $\Q$. We have 
\[
\A^{(j,j+1)} \approx \left(\frac{\nu}{2}\right)^{2(j-1)} \ \W 
\]
and so besides the main diagonal elements 
$Q_{j,j} \approx (\nu/2)^{2(j-1)} \ \|\w\|^2_2$ we have sub- and super-diagonal elements
\[
Q_{j+1,j} = Q_{j,j+1} \approx \left(\frac{\nu}{2}\right)^{2(j-1)} \ \w^\tr \W \w,
\] 
which, depending on $\W$ and $\w$, can be substantially smaller than $Q_{j,j}$. For example, if both $\W$ and $\w$ are generated element-wise independently from zero mean distributions, then for large $N$, 
$\W \w \approx 0$. This is because each row of $\W$ contains i.i.d. realizations of a random variable uncorrelated with random variable whose realizations are stored as elements of $\w$. Then $\w^\tr \W \w$ is negligible.

For elements $Q_{i,j}$ further away from the diagonal, we first analyze properties of the matrix $\B = \W^2$.
\begin{eqnarray}
B_{i,i}
&=& 
\sum_{k=1}^N W_{i,k} \ W_{k,i}
\nonumber \\
&=&
W_{i,i}^2 + \sum_{k \neq i} W_{i,k} \ W_{k,i}
\nonumber \\
&\approx& 
W_{i,i}^2,
\nonumber 
\end{eqnarray}
because of uncorrelated $W_{i,k}$ and $W_{k,i}$ for $k \neq i$. Similarly, for $i \neq j$,
\[
B_{i,j}
=
\sum_{k=1}^N W_{i,k} \ W_{k,j}
\approx 0.
\]
We have  
\begin{eqnarray}
Q_{j+2,j} = Q_{j,j+2} 
&\approx& 
\left(\frac{\nu}{2}\right)^{2(j-1)} \ \w^\tr \ \B \ \w
\nonumber \\
&\approx& 
\left(\frac{\nu}{2}\right)^{2(j-1)} \ \sum_{i=1}^N
W_{i,i}^2 \ w_i^2.
\label{eq:Q_jp2_j}
\end{eqnarray}
Note that in order to scale a large matrix $\til \W$ generated i.i.d. from a zero-mean distribution to spectral radius less than one, the individual elements $W_{i,j}$ of the scaled matrix $\W$ need to be necessarily small, increasingly so for increasing dimensionality $N$. In particular,
based on (\ref{eq:sigma}), the mean of $W_{i,i}^2$ is approximately
$\nu^2/(4 N)$. Comparing (see eq. (\ref{eq:approx_Q_jj}))
\[
Q_{j,j} \approx \left(\frac{\nu}{2}\right)^{2(j-1)} \ \sum_{i=1}^N
w_i^2
\]
with eq. (\ref{eq:Q_jp2_j}), 
we see that there will be an increasing gap (with increasing state space dimensionality $N$) between the diagonal 
elements $Q_{j,j}$ of $\Q$ and the corresponding elements $Q_{j+2,j}=Q_{j,j+2}$ two places off the diagonal.  

Continuing the preceding argumentation inductively, we can conclude that compared to the diagonal terms $Q_{j,j}$ of $\Q$, for the approximation purposes, the off-diagonal terms can be neglected and the metric tensor can be approximated by a diagonal matrix
\begin{equation}
\Q \approx \widehat \Q = \| \w \|^2_2 \ \
\hbox{diag}\!\left(
1, 
\left(\frac{\nu}{2}\right)^{2}, 
\left(\frac{\nu}{2}\right)^{4}, ...,
\left(\frac{\nu}{2}\right)^{2(\tau-1)}
\right).
\label{eq:approx_Q}
\end{equation}

\subsection{Temporal kernel motifs generated by random $\W$}
The eigen-decomposition of $\widehat \Q$ is straightforward:
The eigenvectors form the standard basis $\{ \e_i \}$, each vector $\e_i$ containing zeros, except for the $i$-th element, which is equal to 1.
The corresponding eigenvalues are equal to the diagonal elements of $\widehat \Q$, 
\begin{equation}
\widehat \lambda_i = \| \w \|^2 \ \left(\frac{\nu}{2}\right)^{2(i-1)}.
\label{eq:hat_lambda_i}
\end{equation}
This means that the motif
$\m_i = \e_i$ extracts the $i$-th element from the history of the time series and weights it with the weight $\| \w \| \ (\nu/2)^{i-1}$.

Perhaps surprisingly, the temporal kernel defined by the dynamical system (\ref{eq:state}) with random coupling $\W$ generated i.i.d. from a zero mean distribution has a rigid Markovian flavor with shallow memory. In particular, the kernel
\begin{eqnarray}
K(\u,\v)
&=& 
\sum_{i=1}^N
\lambda_i \ \lla \m_i, \u \rra
\ \lla \m_i, \v \rra
\nonumber \\
&\approx& 
\sum_{i=1}^N
\widehat \lambda_i \ \lla \e_i, \u \rra
\ \lla \e_i, \v \rra
\nonumber \\
&\approx& 
\| \w \|^2 \ \sum_{i=1}^N
 \left(\frac{\nu}{2}\right)^{2(i-1)}
u_i \ v_i,
\nonumber 
\end{eqnarray}
compares the corresponding recent entries of the time series and weights down comparisons of past elements with rapidly decaying weights.

\begin{figure}
\centering
\includegraphics[width=8cm,height=7.5cm]{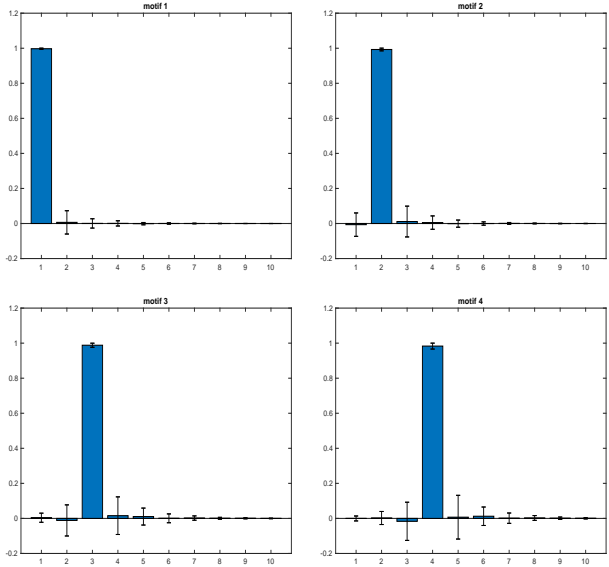}
\caption{The first 10 elements of the four most dominant kernel motifs corresponding to $\W \in \RR^{100 \times 100}$ 
generated element-wise i.i.d. from $N(0,1)$ and renormalized to largest singular value $\nu=0.995$. The input coupling $\w$ was generated element-wise i.i.d. from $N(0,1)$ and renormalized to unit length. Shown are the means and standard deviations across 100 joint realizations of $\W$ and $\w$.  
}
\label{fig:motifs_w_randn}
\end{figure}

\begin{figure}
\centering
\includegraphics[width=5.5cm,height=5cm]{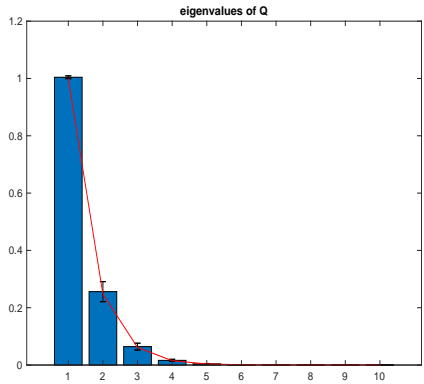}
\caption{Eigenvalues (squared motif weights) of the metric tensor $\Q$ for random setting of the dynamical system (\ref{eq:state}) as described in fig. \ref{fig:motifs_w_randn}. 
Solid bars correspond to the means of the actual eigenvalues $\lambda_i$ across the 100 realizations of $\W$ and $\w$ (also shown are standard deviations). The theoretically predicted values (eq. (\ref{eq:hat_lambda_i})) are shown as the red line. 
}
\label{fig:eigenvalues_w_randn}
\end{figure}

To illustrate this approximation, as well as the rapidly decaying memory of such temporal kernels, we considered 100-dimensional state space ($N=100$) and generated 100 realizations
of $\til \W$ with elements $W_{i,j}$ randomly distributed according to the standard normal distribution $N(0,1)$. Each $\til \W$ was renormalized to 
$\W$ of {largest singular value $\nu=0.995$} and an input coupling vector $\w$ was generated as a random vector with elements generated i.i.d. according to $N(0,1)$ and then renormalized to unit vector (length 1).
We then imposed a past horizon $\tau=200$ and calculated the metric tensor $\Q$, as well as its approximation
$\widehat \Q$ (eq. (\ref{eq:approx_Q})). 
In figure \ref{fig:motifs_w_randn} we show the true motifs $\m_i$ (eigenvectors of $\Q$ for the first four dominant motifs (motifs with the largest 4 motif weights) as the mean and standard deviations across the 100 realizations. For clarity, we only show the first 10 dimensions. It is clear that the motifs approximately correspond to the first four standard basis vectors $\e_i$, $i=1,2,3,4$, as predicted by our theory. Figure \ref{fig:eigenvalues_w_randn} presents the corresponding eigenvalues - solid bars correspond to the means of the actual eigenvalues $\lambda_i$ across the 100 realizations (also shown are standard deviations). The theoretically predicted values (eq. (\ref{eq:hat_lambda_i})) are shown as the red line. Again, there is a strong agreement between the theoretical approximations $\widehat \lambda_i$ and the real eigenvalues $\lambda_i$. 
{We illustrate generality of this result in appendix \ref{sec:appendix}, where motifs and their weights 
were obtained under the same conditions, but with
the input coupling vector $\w$ generated as a vector of all 1s with randomly flipped signs (with equal probability 0.5 in each dimension).
We also tried setting where 
both $\til \W$ and $\w$ consist of all 1s with signs flipped independently element-wise with probability 0.5.
In both cases the Markovian motifs and their weights are almost indistinguishable from those shown in figures \ref{fig:motifs_w_randn}
and \ref{fig:eigenvalues_w_randn}.
}

It is notable that even though the state space dimensionality is quite high ($N=100$), the rapidly decaying motif weights basically prevent the kernel to be able to dig deeper into the history of the time series $\u$, $\v$ when creating a quantitative evaluation of their similarity, $K(\u,\v)$. If the time series are zero-mean, the kernel is estimating a weighted covariance of $\u$ and $\v$ with weights $\left(\frac{\nu}{2}\right)^{2(i-1)}$ exponentially decreasing at the rate much faster than the upper bound $(\nu)^{2(i-1)}$, given the contractive dynamics of (\ref{eq:state}) with spectral radius $\nu$.

{
We conclude this section by noting that for large random $\W$, the spectral radius $\rho \approx \nu/2$. Hence, the resulting temporal kernel can be readily interpreted from the standpoint of spectral radius: 
The Markovian motifs $\e_i$ have weights $\| \w \| \ \rho^{i-1}$, leading to temporal kernel
\begin{equation}
K(\u,\v)
\approx 
\| \w \|^2 \ \sum_{i=1}^N
 \rho^{2(i-1)}
u_i \ v_i.
\nonumber 
\end{equation}
\cite{Zhang2012}
studied echo state networks with i.i.d. random
weights in $\W$. 
They showed that the dynamic mapping (\ref{eq:state}) can be contractive with high probability even when only the spectral radius $\rho$ (as opposed to maximum singular value $\nu$) is less than one.
}

\section{Symmetric dynamic coupling $\W$}
\label{sec:rand_sym_W}
{In this section we investigate how the nature of the temporal kernel changes if we impose symmetry  on the dynamic coupling $\W$ of system (\ref{eq:state}): $W_{i,j}=W_{j,i}$, $i,j = 1,2,...,N$.
In this case, the largest singular value $\nu$ of $\W$ is equal to its spectral radius.
Memory capacity of such systems was rigorously analyzed in \cite{Tino2013,Tino2018_Fisher}. In terms of memory capacity, the role of self-couplings in large systems was shown to be negligible.
In \cite{Couillet2016-SSP} systems with symmetric coupling were shown to lead to inferior performance on memory tasks, when compared with unconstrained dynamic coupling. Similar observation was made in the context of forecasting realized variances of stock market indices \cite{Ficura2017}. }

Recall that given $N_k$ kernels $K^{(a)}(\cdot,\cdot)$  operating on a space $\cX$ and positive real numbers $\alpha_a >0$, $a=1,2,...,N_k$, the linear combination $K(\cdot,\cdot) = \sum_{a=1}^{N_k} \alpha_a K^{(a)}(\cdot,\cdot)$ is a valid kernel on $\cX$. We will show that in case of symmetric $\W$, the corresponding temporal kernel can be understood as a linear combination of simple kernels, each with a unique exponentially decaying motif. 

\vskip 0.5cm
\begin{theorem}
\label{Thm:Symm_W}
Consider the dynamical system (\ref{eq:state}) of state dimensionality $N$
with symmetric coupling $\W$ of rank $N_k \le N$. Let $\s_1, \s_2, ..., \s_{N_k}$, be the eigenvectors of $\W$ corresponding to non-zero eigenvalues $\sigma_1 \ge \sigma_2 \ge ... \ge \sigma_{N_k}$. Denote by $\til w_a = \s_a^\tr \w$ the projection of the input coupling $\w$ onto the eigenvector $\s_a$. Then the temporal kernel $K(\cdot,\cdot)$ associated with system (\ref{eq:state}) is a linear combination of $N_k$ kernels $K^{(a)}(\cdot,\cdot)$,
\begin{equation}
K(\cdot,\cdot) = \sum_{a=1}^{N_k} \til w_a^2 \ K^{(a)}(\cdot,\cdot),
\label{eq:K_symm_W}
\end{equation}
each kernel $K^{(a)}$ with a single motif 
\begin{equation}
\m^{(a)} = (1,\sigma_a,\sigma_a^2,...,\sigma_a^{\tau-1})^\tr \in \RR^\tau.
\label{eq:motif_K^a}
\end{equation}

\end{theorem}

\vskip 0.5cm
\proof
Since $\W$ is symmetric, it can be decomposed as
\[
\W = \S \ \Sigma \ \S^\tr,
\]
where $\S = [\s_1, \s_2, ..., \s_{N_k}]$ is an $N \times N_k$ matrix storing the eigenvectors of $\W$ as columns, with the associated eigenvalues organized along the diagonal of $\Sigma = \hbox{diag}(\sigma_1, \sigma_2, ...,\sigma_{N_k})$. The powers of $\W$ can be then expressed simply through powers of $\Sigma$:
$\W^i = \S \ \Sigma^i \ \S^\tr$.
We thus have
\begin{eqnarray}
Q_{i,j} 
&=&
\w^\tr \ (\W^\tr)^{i-1} \ \W^{j-1} \ \w
\nonumber \\
&=&
\w^\tr \ \W^{i+j-2} \ \w   \hskip 1cm  \hbox{(by symmetry of $\W$)}
\nonumber \\
&=&
\w^\tr \S \ \Sigma^{i+j-2} \ \S^\tr \w 
\nonumber \\
&=&
\til \w^\tr \ \Sigma^{i+j-2} \ \til \w,
\label{eq:Q_ij_W_symm}
\end{eqnarray}
where $\til \w = \S^\tr \w$ is the projection of input coupling $\w$
onto the orthonormal eigen-basis of $\W$.

Writing (\ref{eq:Q_ij_W_symm}) as a quadratic form, we obtain
\begin{eqnarray}
Q_{i,j} 
&=&
\sum_{a,l=1}^{N_k} \til w_a \ \til w_l \ \Sigma^{i+j-2}_{a,l}
\nonumber \\
&=&
\sum_{a=1}^{N_k} 
\til w^2_a \ \sigma^{i+j-2}_a,
\label{eq:Q_ij_W_symm_1}
\end{eqnarray}
because $\Sigma$ is a diagonal matrix.

Let us define $N_k$ matrices $\Q^{(a)} \in \RR^{\tau \times \tau}$, $a=1,2,...,N_k$, as
\[
Q^{(a)}_{i,j} = \sigma^{i+j-2}_a.
\]
Then, 
\begin{equation}
\Q = \sum_{a=1}^{N_k} \til w^2_a \ \Q^{(a)}.
\label{eq:Q_symm_W_a}
\end{equation}

Note that $\Q^{(a)}$ are rank-1 positive semi-definite matrices 
$\Q^{(a)}= \m^{(a)} \ (\m^{(a)})^\tr$. 
Since 
\[
\Q^{(a)} \ \m^{(a)} = \m^{(a)} \ (\m^{(a)})^\tr \ \m^{(a)} =
\| \m^{(a)} \|^2_2 \ \m^{(a)},
\]
we have that $\m^{(a)}$ is the only eigenvector of $\Q^{(a)}$ with a non-zero eigenvalue, i.e. $\m^{(a)}$ is the only motif of the kernel
\[
K^{(a)}(\u,\v) = \u^\tr \ \Q^{(a)} \ \v
\]
with non-zero motif weight.
From (\ref{eq:Q_symm_W_a}) it follows that
$K(\u,\v) = \sum_{a=1}^{N_k} \til w_a^2 \ K^{(a)}(\u,\v)$.
\ep
\vskip 0.5cm

Theorem \ref{Thm:Symm_W} states that the temporal kernel of a system (\ref{eq:state}) with symmetric coupling is a linear combination of several kernels, each of which corresponds to a single non-zero eigenvalue $\sigma_a$ of $\W$. Each such kernel has a unique motif $\m^{(a)}\in \RR^\tau$ associated with it. The motifs $\m^{(a)}$ can only be of two kinds: Either an exponentially decaying profile $(1,\sigma_a,\sigma_a^2,\sigma_a^3,\sigma_a^4, ...)$, if $\sigma_a$ is positive, or an exponentially decaying profile with high oscillation frequency
$(1,-|\sigma_a|,\sigma_a^2, -|\sigma_a^3|, \sigma_a^4,...)$, if $\sigma_a$ is negative. This is obviously quite limiting, precluding the component kernels $K^{(a)}(\cdot,\cdot)$ to capture more diverse range of possible dynamic behaviors.
  
A word of caution is in order. The individual motifs $\m^{(a)}$ are indeed motifs of the component kernels $K^{(a)}(\cdot,\cdot)$, but they are not motifs of the kernel $K(\cdot,\cdot)$. Even though one can write
\[
\Q = \V \ \Sigma_W \ \V^\tr,
\]
where the matrix $\V = [\m^{(1)}, \m^{(2)}, ..., \m^{(N_k)}]$ stores component motifs
$\m^{(a)}$ as columns and 
$\Sigma_W = \hbox{diag}
(\til w_1^2, \til w_a^2, ..., \til w_{N_k}^2)$,
the component motifs $\m^{(a)}$ are not orthogonal. Hence, in general there is no non-zero number $\kappa$, such that
\[
\Q \ \m^{(a)} = \V \ \Sigma_W \ \V^\tr \ \m^{(a)} = \kappa \ \m^{(a)}.
\]
 
Unlike in the previous section, because of the imposed symmetry on $\W$, it is much more difficult to approximate the structure of $\Q$. 
We can recover the upper bound (\ref{eq:abs_Q_ij}) of theorem \ref{Thm:Q} on absolute values of $Q_{i,j}$. From Theorem \ref{Thm:Q} and eq. (\ref{eq:Q_ij_W_symm_1})  we have 
\[
Q_{i,j} 
=
\sum_{a=1}^{N} 
\til w^2_a \ \sigma^{i+j-2}_a 
\]
and 
\begin{eqnarray}
|Q_{i,j}| 
&\le&
\sum_{a=1}^{N} 
\til w^2_a \ |\sigma^{i+j-2}_a| 
\nonumber \\
&\le&
\nu^{i+j-2}_a \sum_{a=1}^{N} \til w^2_a  
\label{eq:Q_i,j_ineq} \\
&\le&
\nu^{i+j-2} \ \| \w \|_2^2.
\label{eq:Q_i,j_eq}
\end{eqnarray}
Here (\ref{eq:Q_i,j_eq}) follows from (\ref{eq:Q_i,j_ineq}) since the norm of the input coupling $\w$ is invariant with respect to orthonormal change of basis.
The inequality in (\ref{eq:Q_i,j_eq}) becomes equality if $\W$ is full rank.

\section{$\W$ as a scaled permutation matrix}
\label{sec:scr_W}

{
We will now consider a strongly constrained dynamical coupling $\W$ in the form of cyclic 
$N \times N$ permutation matrix $\P$, scaled by $\nu$, so that the largest singular value, as well as the spectral radius of
$\W = \nu \cdot \P$ is equal to $\nu$. This follows from a theorem by Frobenius that states that for a non-negative matrix $\W$, its spectral radius is lower and upper bounded by the minimum and maximum row sum, respectively (e.g. \cite{Minc88}). Since in our case all rows of $\W$ sum to $\nu$, the spectral radius must be $\nu$\footnote{
Alternatively, this can be shown by arguing that  $\W$ is a normal matrix.}}

Without loss of generality\footnote{
We can always renumber the state space dimensions.}
we will consider cyclic permutation $\{ 1 \to 2, 2 \to 3, ..., N-1 \to N, N \to
 1 \}$, represented by 
$P_{i+1,i} =1$, $i=1,2,...,N-1$ and $P_{1,N}=1$, all the other elements of $\P$
 are zero.
{Dynamic couplings in the form of scaled cyclic permutation matrix correspond to the setting of simple cycle reservoir \cite{Rodan_Tino2010}, where the reservoir units are connected in a uni-directional ring structure, with the same weight value on all connections in the ring. Analogously, setting of the input coupling $\w$ can be very simple, controlled again by a single amplitude value $w>0$ for all input weights. Intuitively, all the input weights should not have the same value $w$, as this would greatly symmetrize the ESN architecture. To break the symmetry, 
\cite{Rodan_Tino2010} suggest to apply an a-periodic sign pattern to the input weights (e.g. according to binary expansion of an irrational number). While such a reservoir structure has the advantage of being extremely simple and completely deterministic, the predictive performance of the associated ESNs in a variety of tasks on time series of different origins and memory structure was shown to be on par (and sometimes even better) with the usual random reservoir constructions \cite{Rodan_Tino2010}.
Similar observations were made in \cite{Strauss2012}.
This is of great practical importance, since many optics-based physical constructions of reservoir models follow the ring topology structure, which can be implemented using a single unit with multiple delays \cite{Rohm_2018,TANAKA2019,Appeltant2011}. 
Yet, it has been unclear, why such a simple setting can be competitive in real word tasks, or why indeed the breaking of symmetry through a-periodic sign pattern in the input weights is so crucial. In this section, we will study the nature of motifs associated with ring reservoir topologies and the consequences of adopting periodic, rather than a-periodic input weight sign patterns.
}

Given a  
time horizon $\tau = \ell N$, for some positive integer $\ell>1$,
we will now show that the temporal kernel motifs corresponding to the dynamical system (\ref{eq:state}) with scaled permutation coupling $\W = \nu \cdot \P$ have an intricate block structure.

\vskip 0.5cm
\begin{theorem}
\label{Thm:motifs_W_perm}
Consider the dynamical system (\ref{eq:state}) of state space dimensionality $N$,
with coupling $\W=\nu \cdot \P$, where $\nu \in (0,1)$ and $\P$ is the $N \times N$  cyclic permutation matrix. 
Let $\til \m_i \in \RR^N$, $i = 1,2,...,N$, be motifs of the temporal kernel associated with (\ref{eq:state}) under past time horizon  equal to $N$. Denote the corresponding motif weights by $\til \omega_i$.
Then, given a different past time horizon $\tau = \ell\cdot N$, for some positive integer $\ell >1$,
the temporal kernel motifs $\m_i \in \RR^\tau$ associated with (\ref{eq:state}) have the following block form:
\[
\m_i = \left(\til \m_i^\tr, \nu^N \til \m_i^\tr,  \nu^{2 N} \til \m_i^\tr,...,
 \nu^{(\ell-1) N} \til \m_i^\tr\right)^\tr, \ \ \ i=1,2,...N.
\]
The corresponding motif weights are equal to 
\[
\omega_i = \til \omega_i \ 
\left(
\frac{1-\nu^{2 \tau}}{1 - \nu^{2 N}}
\right)^{\frac{1}{2}}.
\]
\end{theorem}

\vskip 0.5cm
\proof
Note that because $\P$ is a permutation matrix, for any non-negative integer $i \in \NN_0$, we have 
\[
\P^i = \P^{N \cdot (i \backslash N)} \ \P^{i \!\!\!
 \hbox{mod} \! N},
\]
where
$\hbox{mod}$  and $\backslash$ denote the modulo and integer division  operations. Since
$\P^{N \cdot (i \backslash N)} = \I_{N \times N}$, we have $\P^i = \P^{i \!\!\! \hbox{mod} \! N}$.
 Consequently,
\[
\W^i = \nu^i \cdot \P^{i \!\!\!\! \hbox{mod} \! N}.
\]
Furthermore, since $\P$ is orthogonal, $\P^{-1}=\P^\tr$.
We can now write the elements of $\Q$ as (see eq. (\ref{eq:Q_ij})),
\begin{eqnarray}
Q_{i,j} 
&=&
\w^\tr \ (\W^\tr)^{i-1} \ \W^{j-1} \ \w
\nonumber \\
&=&
\nu^{i+j-2} \ \w^\tr \ (\P^\tr)^{i-1} \ \P^{j-1} \ \w
\nonumber \\
&=&
\nu^{i+j-2} \ \w^\tr \ \P^{j-i} \ \w.
\nonumber \\
&=&
\nu^{i+j-2} \ \w^\tr \ \P^{(j-i) \!\!\!\! \hbox{mod} \! N} \ \w.
\label{eq:Q_ij_W_perm}
\end{eqnarray}

For $k \in \{-N+1, -N+2,...,-1,0,1,...N-1 \}$, if $k$ is positive, $\P^k  \w$ is
 the vector with elements of $\w$ rotated $k$ places to the right. In case $k$ 
is negative, the rotation is to the left. 
From (\ref{eq:Q_ij_W_perm}) it is clear the $\Q \in \RR^{\tau \times \tau}$ has the following block structure.
\[
\Q=
\left[
\begin{array}{c c c c}
\Q^{(1,1)} & \Q^{(1,2)} & \cdots & \Q^{(1,\ell)} \\
\Q^{(2,1)} & \Q^{(2,2)} & \cdots & \Q^{(2,\ell)} \\
\cdots & \cdots & \cdots & \cdots \\
\Q^{(\ell,1)} & \Q^{(\ell,2)} & \cdots & \Q^{(\ell,\ell)} 
\end{array}
\right],
\]
where each matrix $\Q^{(a,b)} \in \RR^{N \times N}$, $a,b = 1,2,..,,\ell$, has elements
\[
Q^{(a,b)}_{i,j} = \nu^{(a+b-2) N} \ \nu^{i+j-2} \ \w^\tr \ \P^{j-i} \ \w,
\ \ \ i,j=1,2,...,N.
\]

Define an ${N \times N}$ matrix $\R$ with elements
\begin{equation}
R_{i,j} =  \nu^{i+j-2} \ \w^\tr \ \P^{j-i} \ \w,
\ \ \ i,j=1,2,...,N,
\label{eq:R}
\end{equation}
yielding  $\Q^{(a,b)} = \nu^{(a+b-2) N} \ \R$.
Note that $\R$ is the metric tensor of the temporal kernel associated with (\ref{eq:state}) under the past time horizon $N$. Let $\til \m_i \in \RR^N$ be the $i$-th eigenvector of $\R$ with eigenvalue $\til \lambda_i$. Then,
\[
\Q^{(a,b)} \ \til \m_i
= \nu^{(a+b-2) N}\ \til \lambda_i \ \til \m_i,
\]
and so
\begin{equation}
\left[ \Q^{(a,1)}, \Q^{(a,2)}, \cdots, \Q^{(a,\ell)} \right]
\left[
\begin{array}{c}
\til{\m_i} \\
\til \m_i \\
\cdots\\
\til \m_i 
\end{array}
\right]
=
\til \lambda_i \ 
\left(
\sum_{j=1}^\ell
\nu^{(j-1) N}
\right)
\nu^{(a-1) N} \
\ \til{\m_i}. 
\end{equation}
It follows that for each $a \in \{1,2,...,\ell\}$,
\begin{equation}
\left[ \Q^{(a,1)}, \Q^{(a,2)}, \cdots, \Q^{(a,\ell)} \right]
\left[
\begin{array}{r}
\til{\m_i} \\
\nu^N \ \til \m_i \\
\cdots\\
\nu^{(\ell-1) N}\ \til \m_i 
\end{array}
\right]
=
\til  \lambda_i \ 
\left(
\sum_{j=1}^\ell
\nu^{2 (j-1) N}
\right)
\nu^{(a-1) N} \
\ \til{\m_i}. 
\end{equation}
We can thus conclude that the vector
\[
\m_i = \left(\til \m_i^\tr, \nu^N \til \m_i^\tr,  \nu^{2 N} \til \m_i^\tr,...,
 \nu^{(\ell-1) N} \til \m_i^\tr\right)^\tr
\]
is an eigenvector of $\Q$ with eigenvalue
\begin{eqnarray}
\lambda_i 
&=& 
\til  \lambda_i \ 
\sum_{j=0}^{\ell-1}
\left( 
\nu^{2 N}
\right)^j 
\nonumber \\
&=& 
\til  \lambda_i \ 
\frac{1-\nu^{2 N \ell}}{1 - \nu^{2 N}}. 
\nonumber \\
&=& 
\til  \lambda_i \ 
\frac{1-\nu^{2 \tau}}{1 - \nu^{2 N}}. 
\nonumber 
\end{eqnarray}

\ep
\vskip 0.5cm

\subsection{Periodic input coupling $\w$}
It has been empirically shown in \cite{Rodan_Tino2010} that when the dynamic coupling $\W$ is formed by a scaled permutation matrix, a very simple setting of input coupling $\w$ is sufficient: all elements of $\w$ can have the same absolute value, but the sign pattern should be aperiodic.
Intuitively, it is clear that for such $\W$ a periodic input coupling $\w$ will induce symmetry in the dynamic processing of (\ref{eq:state}) and such a symmetry should be broken. However, in this section we would like to ask exactly what representational capabilities are lost by imposing a periodicity in $\w$.

We will start by considering a general case of periodic $\w \in \RR^N$ formed by $k>1$ copies of a periodic block $\s \in \RR^p$,
$\w = (\s^\tr, \s^\tr, ..., \s^\tr)^\tr$. Obviously, $N = k \cdot p$. 

Denote by $\ov\P \in \RR^{p \times p}$ the top left $p \times p$ block of the right shift permutation matrix $\P \in \RR^{N \times N}$. In other words, 
$\ov\P$ is the right shift permutation matrix operating on vectors from $\RR^p$. Furthermore, we introduce matrix $\T \in \RR^{p \times p}$ with elements
\begin{equation}
T_{i,j} = \nu^{i+j-2} \lla \s, \ov\P^{|j-i|} \s \rra, \ \ \ i,j=1,2,...,p.
\label{eq:T}
\end{equation}

\vskip 0.5cm
\begin{theorem}
\label{Thm:motifs_W_perm_w_per}
Consider the dynamical system (\ref{eq:state}) of state space dimensionality $N$,
with coupling $\W=\nu \cdot \P$, where $\nu \in (0,1)$ and $\P$ is the $N \times N$  cyclic permutation matrix. 
Let the input coupling $\w \in \RR^N$ consist of $k>1$ copies of a periodic block $\s \in \RR^p$.
Denote by $\ov\m_i \in \RR^p$, $i = 1,2,...,p$, eigenvectors of the matrix
$\T$ (\ref{eq:T}) with the corresponding eigenvalues $\ov\lambda_i$.
Then, given a past time horizon $\tau = \ell\cdot N$, for some positive integer $\ell >1$,
there are at most $p$ temporal kernel motifs $\m_i \in \RR^\tau$ associated with (\ref{eq:state}) of non-zero motif weight. Furthermore, the kernel motifs have the following block form,
\[
\m_i = (\ov\m_i^\tr, \nu^p \ \ov\m_i^\tr, \nu^{2p} \ \ov\m_i^\tr, ...,
 \nu^{\tau-p} \ \ov\m_i^\tr)^\tr, \ \ \ i=1,2,...p,
\]
with the corresponding motif weights
\[
\omega_i = 
\left(
\ov\lambda_i \ \frac{1-\nu^{2 \tau}}{1 - \nu^{2 p}}
\right)^{\frac{1}{2}}.
\]
\end{theorem}

\vskip 0.5cm
\proof
By Theorem \ref{Thm:motifs_W_perm}, to determine motifs of 
the temporal kernel associated with (\ref{eq:state}),  
it is sufficient to perform eigen-analysis of the block matrix 
$\Q^{(1,1)}=\R$ (eq. (\ref{eq:R})).

For $a=0,1,2,...,N-1$,
\[
\lla \w, \P^{a} \w \rra 
=
k \cdot \lla \s, \ov\P^{\ \! a} \s \rra 
= k \cdot \lla \s, \ov\P^{\ \! a \!\!\!\! \hbox{mod} \! p} \ \s \rra
\]
and since $\R$ is symmetric,
from (eq. (\ref{eq:R})) we have
\begin{equation}
Q^{(1,1)}_{i,j} = k \cdot \nu^{i+j-2} \cdot 
\lla \s, \ov\P^{\ \! |j-i| \!\!\!\! \hbox{mod} \! p} \ \s \rra,
\ \ \ i,j=1,2,...,N.
\label{eq:R_w_per}
\end{equation}
Therefore, 
$\Q^{(1,1)}$ can be decomposed into blocks of $p \times p$ matrices
\[
\Q^{(1,1)}=
\left[
\begin{array}{c c c c}
\C^{(1,1)} & \C^{(1,2)} & \cdots & \C^{(1,k)} \\
\C^{(2,1)} & \C^{(2,2)} & \cdots & \C^{(2,k)} \\
\cdots & \cdots & \cdots & \cdots \\
\C^{k,1)} & \C^{(k,2)} & \cdots & \C^{(k,k)} 
\end{array}
\right],
\]
where
\[
\C^{(c,d)} = \nu^{(c+d-2)p} \ \C^{(1,1)}, \ \ \ \ c,d=1,2,...,k
\]
and
\[
C^{(1,1)}_{i,j} = \nu^{i+j-2} \cdot 
\lla \s, \ov\P^{\ \! |j-i|} \ \s \rra,
\ \ \ i,j=1,2,...,p.
\]

Now, let $\ov\m_i \in \RR^p$ be the $i$-th eigenvector of $\C^{(1,1)} = \T$ with eigenvalue $\ov\lambda_i$. Then,
\[
\C^{(c,d)} \ \ov\m_i
= \nu^{(c+d-2) p}\ \C^{(1,1)} \ \ov\m_i =
\nu^{(c+d-2) p}\ \ov\lambda_i \ \ov\m_i.
\]
We have
\begin{equation}
\left[ \C^{(c,1)}, \C^{(c,2)}, \cdots, \C^{(c,k)} \right]
\left[
\begin{array}{r}
\ov\m_i \\
\nu^p \ \ov\m_i \\
\cdots\\
\nu^{(k-1) p}\ \ov\m_i 
\end{array}
\right]
=
\ov\lambda_i \ 
\left(
\sum_{j=1}^k
\nu^{2 (j-1) p}
\right)
\nu^{(c-1) p} \
\ \ov\m_i 
\end{equation}
for $c =1,2,...,k$.
Hence,
\[
\til \m_i = (\ov\m_i^\tr, \nu^p \ \ov\m_i^\tr,  \nu^{2 p} \ \ov\m_i^\tr,...,
 \nu^{(k-1) p}\ \ov\m_i^\tr)^\tr
\]
is an eigenvector of $\Q^{(1,1)}$ with eigenvalue
\begin{eqnarray}
\til \lambda_i 
&=& 
\ov\lambda_i \ 
\sum_{j=0}^{k-1}
\left( 
\nu^{2 p}
\right)^j 
\nonumber \\
&=& 
\ov\lambda_i \ 
\frac{1-\nu^{2 p k}}{1 - \nu^{2 p}}. 
\nonumber \\
&=& 
\ov\lambda_i \ 
\frac{1-\nu^{2 N}}{1 - \nu^{2 p}}. 
\nonumber 
\end{eqnarray}

By Theorem \ref{Thm:motifs_W_perm}, the corresponding eigenvector $\m_i$ of $\Q$ reads:
\begin{eqnarray}
\m_i 
&=&
(\til \m_i^\tr, \nu^N \til \m_i^\tr, ...,
 \nu^{(\ell-1) N} \til \m_i^\tr)^\tr
\nonumber \\
&=&
(\ov\m_i^\tr, \nu^p \ \ov\m_i^\tr, ..., \nu^{(k-1) p}\ \ov\m_i^\tr,
\nu^N \ \ov\m_i^\tr, \nu^{N+p} \ \ov\m_i^\tr, ..., 
\nu^{(\ell-1) N + (k-1) p}\ \ov\m_i^\tr
)^\tr
\nonumber \\
&=&
(\ov\m_i^\tr, \nu^p \ \ov\m_i^\tr, \nu^{2 p}\ ..., \nu^{\tau - p}\ \ov\m_i^\tr
)^\tr.
\nonumber
\end{eqnarray}
The last equality holds since from $\tau = \ell N$ and $N = k p$, we have
$(\ell-1) N + (k-1) p = \tau - p$.
We can calculate the corresponding eigenvalue as
\begin{eqnarray}
\lambda_i 
&=& 
\til  \lambda_i \ 
\frac{1-\nu^{2 \tau}}{1 - \nu^{2 N}} 
\nonumber \\
&=&
\ov\lambda_i \ 
\frac{1-\nu^{2 N}}{1 - \nu^{2 p}}
\ \frac{1-\nu^{2 \tau}}{1 - \nu^{2 N}}
=
\ov\lambda_i \ 
\frac{1-\nu^{2 \tau}}{1 - \nu^{2 p}}.
\nonumber
\end{eqnarray}

\ep

Theorem \ref{Thm:motifs_W_perm_w_per} formally specifies consequences for the dynamical kernel of having a periodic input coupling $\w$ of period $p$ in the system
(\ref{eq:state}). First, the number of potentially useful kernel motifs of non-zero {weight} is reduced from $N$ (the state space dimensionality) to $p$. Second, the motif structure is even more restricted than in the case of general $\w$. 
If the past horizon is $\tau = \ell N$, then in general, by theorem \ref{Thm:motifs_W_perm}, each motif 
$\m_i \in \RR^\tau$ consists of a series of $\ell$ copies of the same ``core motif" $\til\m_i \in \RR^N$, down-weighted by exponential decay. In the case of periodic $\w$, motifs $\m_i \in \RR^\tau$ are formed by a series
of $\ell k$ copies of the same small block $\ov\m_i \in \RR^p$, down-weighted by exponential decay.

We will now investigate special settings of the periodic input coupling $\w \in \RR^N$. Consider first the binary setting, i.e. the core periodic block is
$\s = (1,0,0,...,0)^\tr \in \{0,1\}^p$. Assume $\w$ contains $k$ such blocks ($N = k \cdot p$).
Then, since for $a = 0,1,2,...,p-1$,
\[
\lla \s, \ov\P^{\ \! a} \ \s \rra
=
\left\{ \begin{array}{rl}
1, & \mbox{if}
\ \ a=0 \\ 
0, & \mbox{otherwise},
\end{array}\right.
\]
the matrix $\T \in \RR^{p \times p}$ (eq. (\ref{eq:T}))
will have a diagonal form,
$\T = \hbox{diag}(1, \nu^2, ..., \nu^{2(p-1)})$.
The eigenvectors $\ov\m_i \in \RR^p$ of $\T$, $i=1,2,...,p$, correspond to the standard basis
$\ov\e_i$ of $\RR^p$, i.e. all elements of $\ov\e_i$ are zeros, except for the $i$-th element, which is 1. The corresponding eigenvalues are $\ov\lambda_i = \nu^{2(i-1)}$.
By theorem \ref{Thm:motifs_W_perm_w_per},
each motif 
\begin{equation}
\m_i  =
(\ov\e_i^\tr, \nu^p \ \ov\e_i^\tr, \nu^{2p} \ \ov\e_i^\tr, ...,
 \nu^{\tau-p} \ \ov\e_i^\tr)^\tr,
\label{eq:m_i_w_per_bin}
\end{equation}
with motif weight
\begin{equation}
\omega_i = 
\nu^{i-1}
\left(
\frac{1-\nu^{2 \tau}}{1 - \nu^{2 p}}
\right)^{\frac{1}{2}}.
\label{eq:m_w_w_bin_per}
\end{equation}
is a periodic exponentially decaying motif that picks up elements of time series driving (\ref{eq:state}) with periodicity $p$ and initial lag $i$.
Given a time series $\u \in \RR^\tau$,
\[
\lla \m_i, \u \rra = 
\sum_{j=1}^{\ell \cdot k}
\nu^{(j-1)p} \ u_{i + (j-1) p}.
\]
In the representation of (eq. (\ref{eq:tilde_u})) we then have
\[
\til\u = 
\left(
\frac{1-\nu^{2 \tau}}{1 - \nu^{2 p}}
\right)^{\frac{1}{2}} 
\cdot
(\lla \m_1, \u \rra, 
\nu \ \lla \m_2, \u \rra, ...,
\nu^{p-1} \ \lla \m_p, \u \rra)^\tr \in \RR^{p}.
\]
Given another time series $\v \in \RR^\tau$, the temporal kernel gives
\begin{eqnarray}
K(\u,\v)
&=&
\lla \til\u, \til\v \rra
\nonumber \\
&=&
\frac{1-\nu^{2 \tau}}{1 - \nu^{2 p}}
\sum_{i=1}^p \nu^{2(i-1)} \ 
\lla \m_i, \u \rra
\
\lla \m_i, \v \rra.
\end{eqnarray}

In the case of all-ones $\w$ with a periodic sign pattern, the
core periodic block is $\s = (+1,-1,-1,...,-1)^\tr \in \{-1,+1\}^p$.
For $a = 0,1,2,...,p-1$, we have
\[
\lla \s, \ov\P^{\ \! a} \ \s \rra
=
\left\{ \begin{array}{ll}
p, & \mbox{if}
\ \ a=0 \\ 
p-4, & \mbox{otherwise},
\end{array}\right.
\]
From (eq. (\ref{eq:T})), the matrix
$\T \in \RR^{p \times p}$ with elements
\[
T_{i,j}
=
\left\{ \begin{array}{ll}
\nu^{2(i-1)} \ p , & \mbox{if}
\ \ i=j \\ 
\nu^{i+j-2} \ (p-4), & \mbox{otherwise},
\end{array}\right.
\]
can yield a richer set of eigenvectors $\ov\m_i$ than the standard basis
$\ov\e_i$ in $\RR^p$. An exception is the case of period-4 sign pattern, $p=4$. In that case, $\T$ is a diagonal matrix
$\T = p \cdot \hbox{diag}(1, \nu^2, ..., \nu^{2(p-1)})$, exactly the scaled version of $\T$ analyzed above, when $\w$ was the binary vector composed of a series of $k$ blocks of $\ov\e_1 \in \RR^p$. Hence the four motifs
$\m_i$ will have the form suggested by eq. (\ref{eq:m_i_w_per_bin}) and the motif weights (\ref{eq:m_w_w_bin_per}) will be scaled by $\sqrt{p}=2$.
We have thus established:

\vskip 0.5cm
\begin{corollary}
\label{Col:motifs_W_perm_w_bin_per}
Under the assumptions of Theorem \ref{Thm:motifs_W_perm_w_per}, assume that
the input coupling $\w \in \{0,1\}^N$ consists of $k>1$ copies of the binary standard basis block $\s = \ov\e_1 \in \{0,1\}^p$.
Then there are $p$ non-zero wight motifs of the dynamic kernel associated with 
(\ref{eq:state}),
\begin{equation}
\m_i  =
(\ov\e_i^\tr, \nu^p \ \ov\e_i^\tr, \nu^{2p} \ \ov\e_i^\tr, ...,
 \nu^{\tau-p} \ \ov\e_i^\tr)^\tr,
\nonumber
\end{equation}
with motif weights
\[
\omega_i = 
\nu^{i-1}
\left(
\frac{1-\nu^{2 \tau}}{1 - \nu^{2 p}}
\right)^{\frac{1}{2}}.
\]
Each $\m_i$ is thus a periodic exponentially decaying motif that picks up elements of input time series with periodicity $p$ and initial lag $i$.

Furthermore, if the bipolar input coupling $\w \in \{-1,+1\}^N$ consists of $k>1$ copies of the block $\s = 2\ \! \ov\e_1 - {\mathbf 1} \in \{-1,+1\}^4$ of period $p=4$, then there are four non-zero wight motifs
$\m_i$ (\ref{eq:m_i_w_per_bin}) with motif weights $2 \omega_i$.
\end{corollary}

\section{Illustrative examples}
\label{sec:examples}
In this section we will illustrate the results obtained so far showing the influence of the dynamic and input coupling, $\W$ and $\w$, respectively, on the strength and richness of motifs of the temporal kernel associated with 
the dynamical system (\ref{eq:state}). In all illustrations we
will use state space dimensionality $N=100$ and {re-normalize the dynamic coupling $\W \in \RR^{100 \times 100}$ to largest singular value $\nu=0.995$}. The input coupling $\w$ is renormalized to unit length. 
{The past horizon will be set to $\tau=200$.}
We will show motifs with motif weights up to $10^{-2}$ of the highest motif weight.

\begin{figure}[ht]
\centering
\includegraphics[width=10cm,height=6cm]{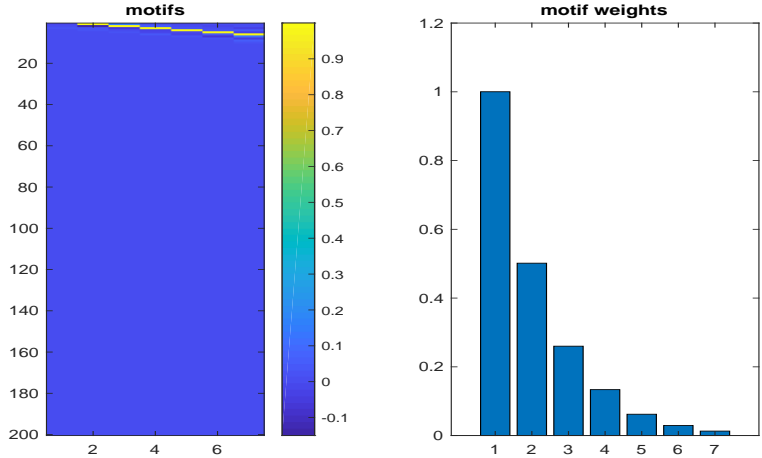}
\caption{Temporal kernel motifs and the corresponding motif weights for randomly generated $\W$ and $\w$.
}
\label{fig:motifs_rand_W_rand_w}
\end{figure}

{
Figure \ref{fig:motifs_rand_W_rand_w} (left) shows motifs of the temporal kernel given by random coupling $\W$, where all elements $W_{i,j}$ were generated i.i.d. from normal distribution $N(0,1)$. 
The motifs are shown in a column-wise fashion, i.e. the x-axis indexes the individual motifs, while the motif values are shown along the y-axis. The
associated motif weights are presented in the right plot.}

As explained in section \ref{sec:rand_W}, each of the Markovian motifs picks an element from the recent time series history, yielding a shallow memory involved in the kernel evaluation, with rapidly decaying motif weights. Almost identical results were obtained for 
$W_{i,j}$ and $w_i$ generated i.i.d. from other distributions (e.g. uniform over $[-1,+1]$, Bernoulli over $\{-1,+1\}$ or $\{0,1\}$), as well as for many other settings of $\w$, including the all-ones vector $\w = {\mathbf 1}$.

\begin{figure}[ht]
\centering
\includegraphics[width=10cm,height=6cm]{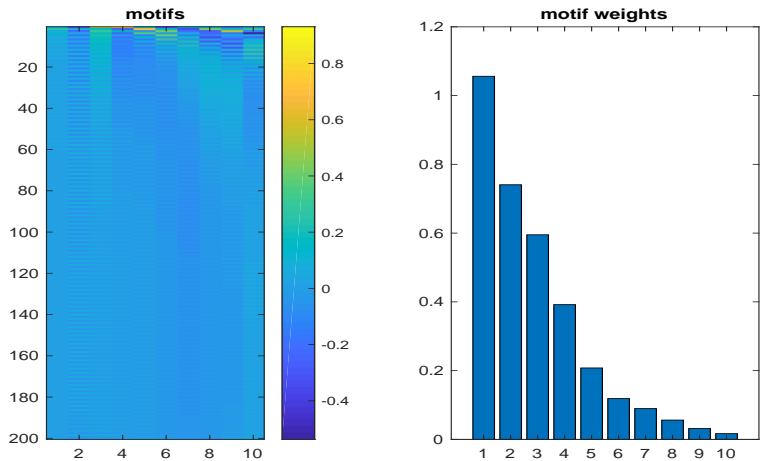}
\caption{Temporal kernel motifs and the corresponding motif weights for random symmetric Wigner $\W$ and random $\w$.
}
\label{fig:motifs_rand_sym_W_rand_w}
\end{figure}

Introduction of a structure into random $\W$ by imposing symmetry (Wigner $\W$) leads to a slightly richer motif set, albeit still with shallow memory
(see figure \ref{fig:motifs_rand_sym_W_rand_w}). Note the high frequency nature of some motifs, as discussed in section 
\ref{sec:rand_sym_W}. Again, the number and nature of the motifs stayed unchanged across a variety of generative mechanisms for $\W$ and $\w$ described above.

\begin{figure}[ht]
\centering
\includegraphics[width=10cm,height=6cm]{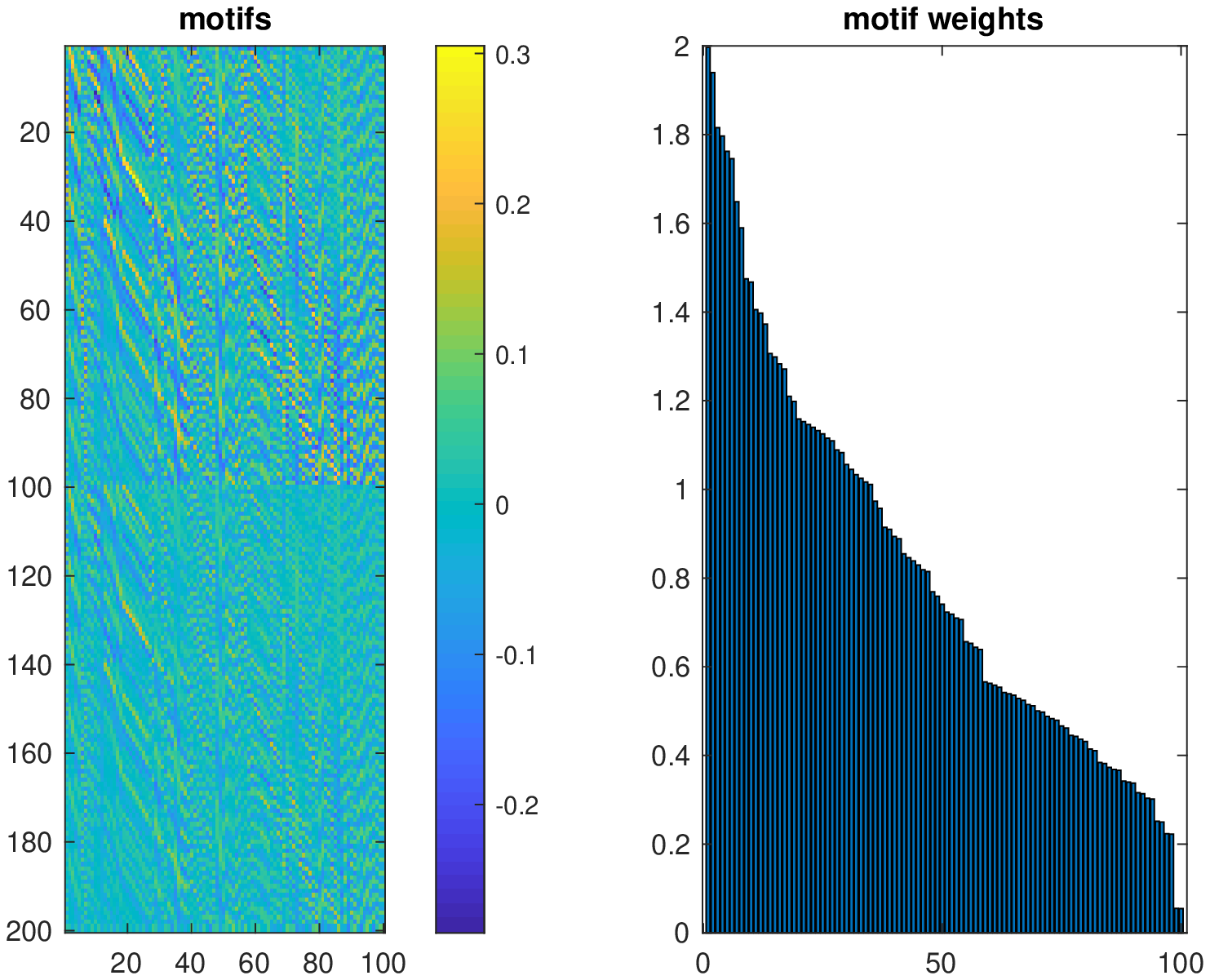}
\caption{Temporal kernel motifs and the corresponding motif weights for scaled permutation matrix $\W$ and random $\w$.
}
\label{fig:motifs_scr_W_rand_w}
\end{figure}

\begin{figure}[ht]
\centering
\includegraphics[width=10cm,height=6cm]{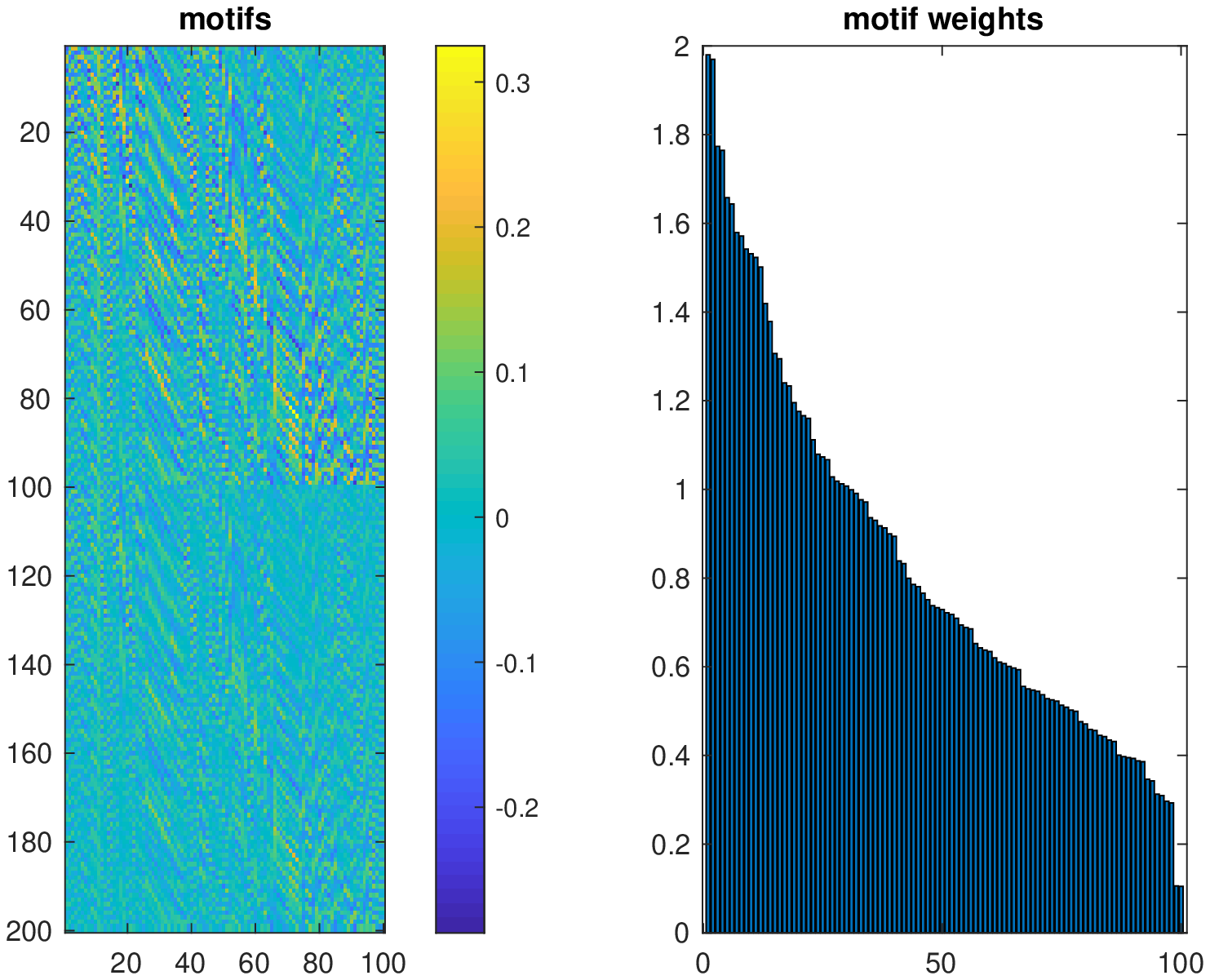}
\caption{Temporal kernel motifs and the corresponding motif weights for scaled permutation matrix $\W$ and aperiodic all-ones vector $\w$ with signs following binary expansion of $\pi$.
}
\label{fig:motifs_scr_W_pi_sign_w}
\end{figure}

\begin{figure}[ht]
\centering
\includegraphics[width=10cm,height=6cm]{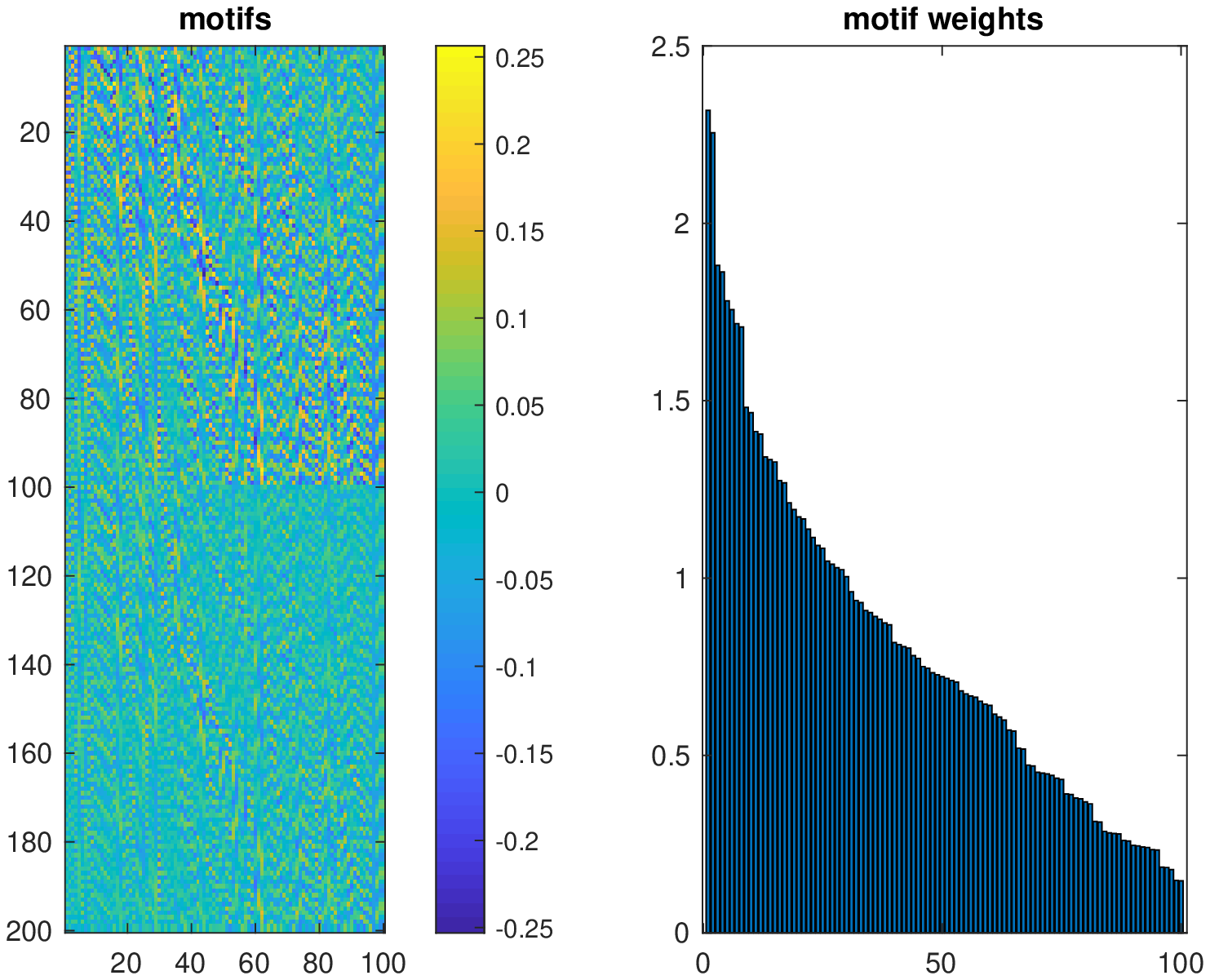}
\caption{Temporal kernel motifs and the corresponding motif weights for scaled permutation matrix $\W$ and aperiodic all-ones vector $\w$ with signs following binary expansion of $e$.
}
\label{fig:motifs_scr_W_e_sign_w}
\end{figure}

\begin{figure}[ht]
\centering
\includegraphics[width=10cm,height=7cm]{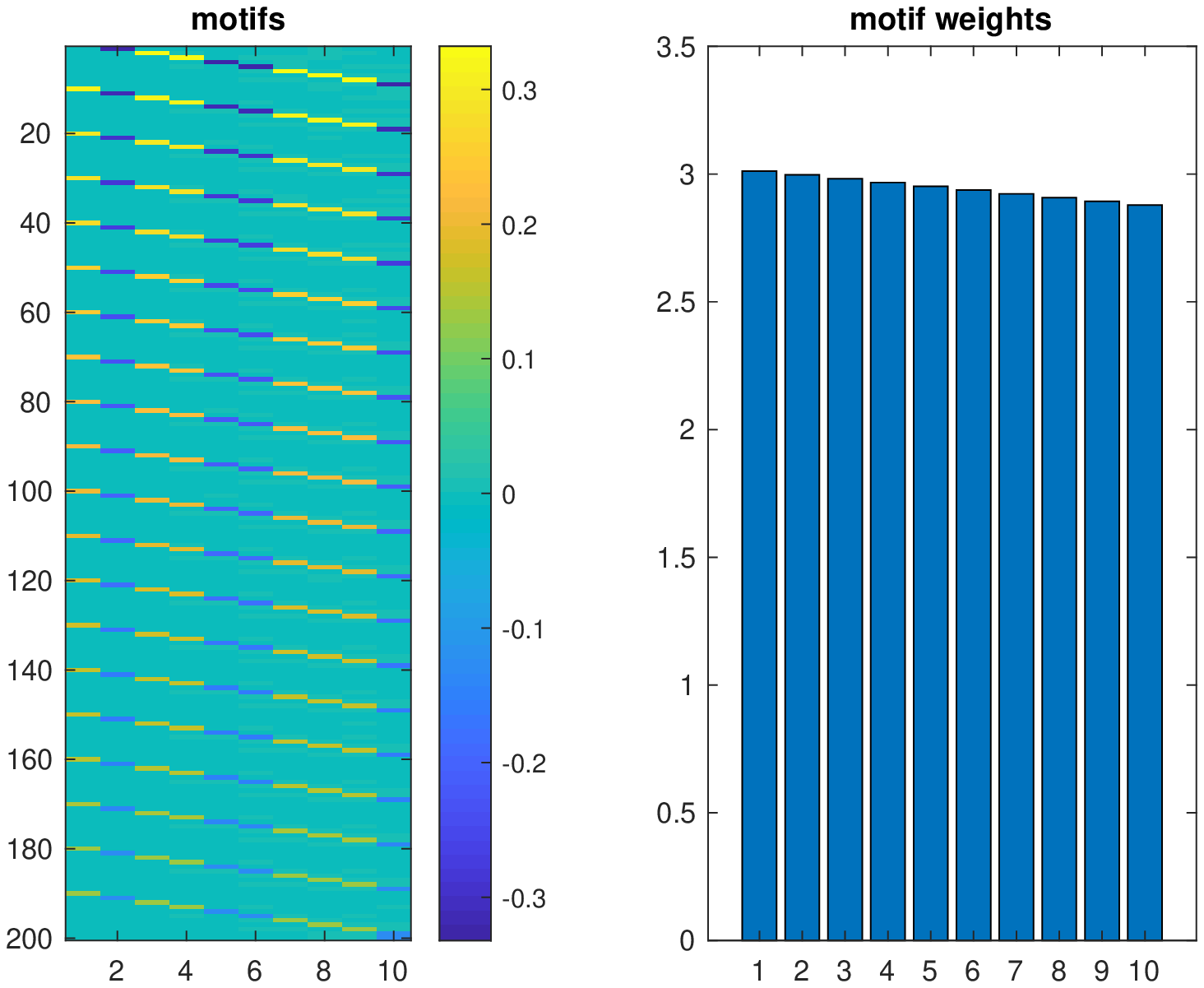}
\caption{Temporal kernel motifs and the corresponding motif weights for scaled permutation matrix $\W$ and periodic binary vector $\w$ with period $p=10$.
}
\label{fig:motifs_scr_W_bin_per_w}
\end{figure}

The situation changes dramatically when $\W$ is set to the scaled permutation matrix
of section \ref{sec:scr_W}.
Figure \ref{fig:motifs_scr_W_rand_w} shows an example of motif and motif weight structure for $\w$ generated randomly i.i.d. from $N(0,1)$. To demonstrate that what really matters, as argued in section \ref{sec:scr_W},
is the aperiodicity of $\w$, we show in figures \ref{fig:motifs_scr_W_pi_sign_w} and \ref{fig:motifs_scr_W_e_sign_w} motifs and motif weights when $\w$ is simply a vector of ones with signs prescribed by the first $N=100$ digits of binary expansion of $\pi$ and $e$, respectively. This was suggested in \cite{Rodan_Tino2010} as a simple controlled way of generating aperiodic input couplings. Such settings admit a full set of $N=100$ highly variable motifs. The scaled block structure of motifs proved in section \ref{sec:scr_W} is clearly visible.
In striking contrast, as suggested in section \ref{sec:scr_W}, we present in figure \ref{fig:motifs_scr_W_bin_per_w} motifs and motif weights for the case of periodic $\w$ with period $p=10$. As predicted by the theory, the shrunk motif set contains $p=10$ simple periodic motifs given by repeated blocks of permuted standard basis $\ov\e_i$ (with possibly flipped sign).

\begin{figure}[ht]
\centering
\includegraphics[width=8cm,height=8cm]{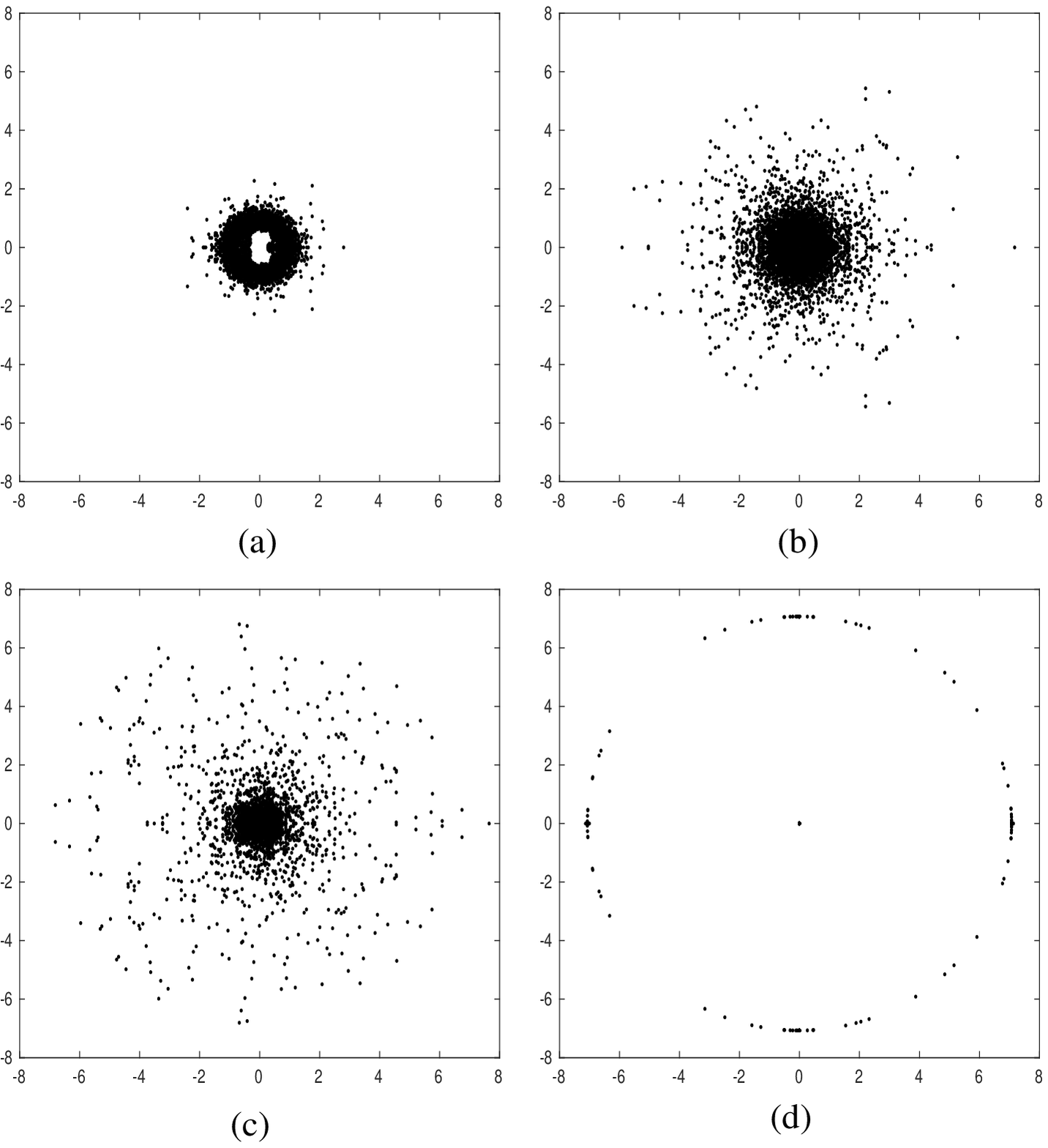}
\caption{Fourier coefficient distribution of motifs of the temporal kernel associated with dynamical system  (\ref{eq:state}) of state space dimensionality $N=100$. Dynamic coupling $\W$ was the scaled permutation matrix with spectral radius $\nu=0.96$ (a), $0.99$ (b), $0.996$ (c) and $1.0$ (d).
The input coupling $\w$ was formed by vector of all $1$s with signs distributed according to the first $N=100$ digits of the binary expansion of $\pi$.
Motifs used have motif weights from the maximal motif weight $\omega_{max}$ to $10^{-2} \omega_{max}$.}
\label{fig:F_coef_motifs}
\end{figure}

In order to quantify ``motif richness", we perform Fast Fourier Transform (FFT) of motifs $\m_i$ with motif weights $\omega_i$ up to $10^{-2}$ of the highest motif weight. 
We collect the Fourier coefficients $z_{i,k} \in \CC$ of each motif $\m_i$ along with the corresponding motif weight $q_{i,k} = \omega_i$ in a set $F_i = \{(z_{i,k},q_{i,k})\}_k$. The total coefficient set is then
$F = \bigcup_i F_i$. 
Figure \ref{fig:F_coef_motifs} presents distribution of motif Fourier coefficients from $F$ for different settings of spectral radius $\nu
\in \{ 0.96, 0.99, 0.996, 1.0 \}$.


\begin{figure}
\centering
\includegraphics[width=10cm,height=7cm]{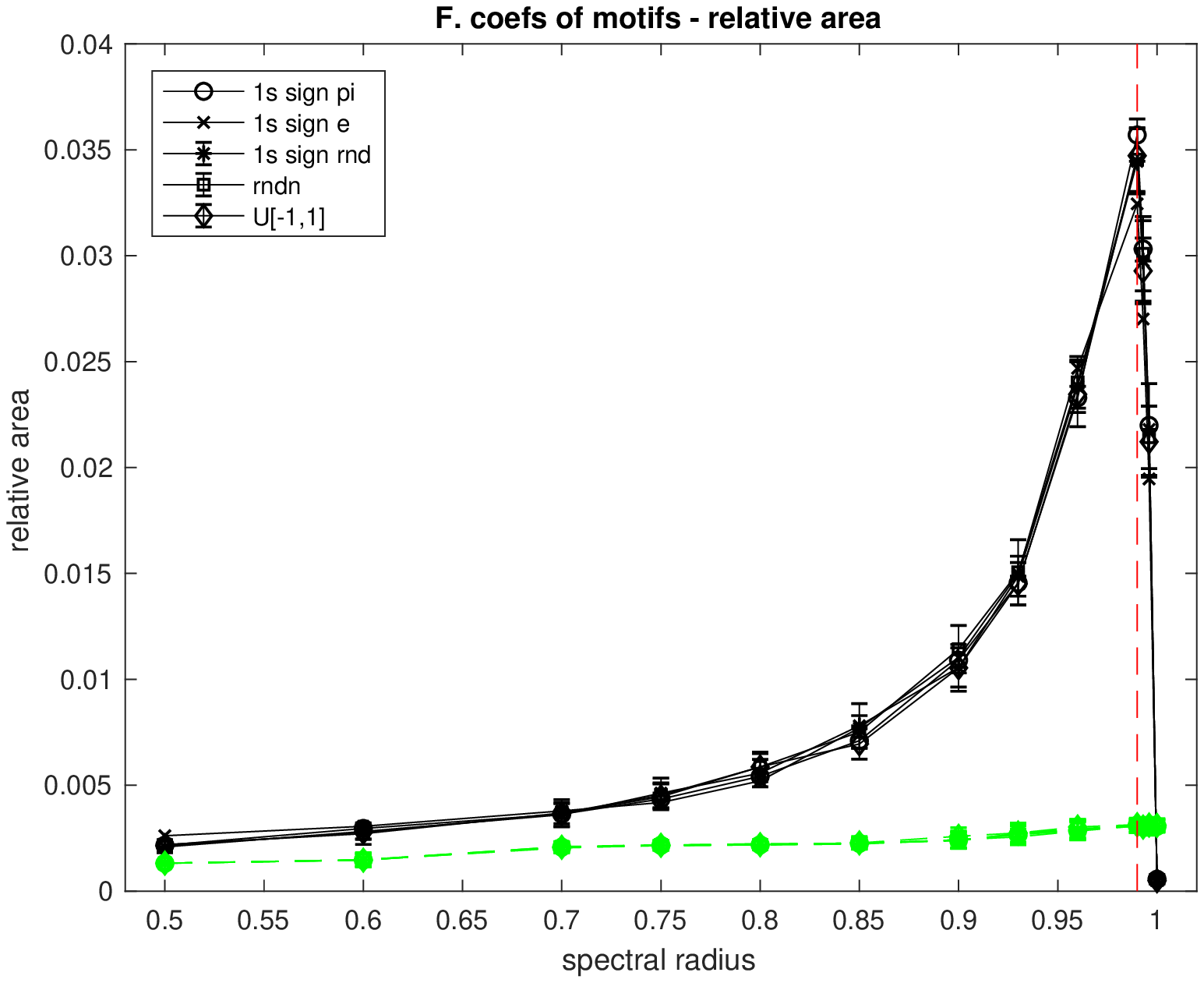}
\caption{Relative area covered by Fourier coefficients of the temporal kernel motifs as a function of largest singular value of $\W$. Dimensionality of the dynamical system  (\ref{eq:state}) was set to $N=100$ with dynamic coupling $\W$ formed by the scaled permutation matrix (black lines) or random matrix with individual elements generated i.i.d. from $N(0,1)$ (dashed green lines).
The input coupling $\w$ was formed by vector of all $1$s with signs distributed randomly (stars), according to the first $N=100$ digits of the binary expansion of $\pi$ (circles) and $e$ (crosses). We also show the case where the elements of $\w$ were generated i.i.d. from $N(0,1)$ (squares) and uniform distribution over $[-1,+1]$ (diamonds). The vectors $\w$ were renormalized to unit length. In case of deterministic $\w$ but stochastic generation of $\W$, the experiment was repeated 30 times. When both $\W$ and $\w$ were generated randomly, the experiment was repeated 60 times. In those cases, we show the means and standard deviations of the relative area covered by the Fourier coefficients.
In each experimental run, the motifs used have their weights ranging from the maximal motif weight $\omega_{max}$ to $10^{-2} \omega_{max}$.}
\label{fig:F_coef_motifs_nu_mean}
\end{figure}

We designed two measures to characterize distribution of Fourier coefficients from $F$. The first is simply the area occupied by the coefficients $z_{i,k}$. In particular,
the coefficient space $[-7,7]^2$  in the complex plane was
covered by regular grid of cells with side length $0.05$.  The {\em relative area} covered by $F$ is then the ratio of the number of cells visited by the coefficients $z_{i,k}$ to the total number of cells.
Figure \ref{fig:F_coef_motifs_nu_mean} shows the relative area occupied by motif Fourier coefficients, {as a function of the scaling largest singular value $\nu$}. It is remarkable how stable the behavior of the relative area is for the scaled permutation matrix $\W$ (black lines), as long as the input coupling $\w$ is aperiodic: we tried 
vector of all $1$s with signs distributed randomly (stars), according to the first $N$ digits of the binary expansion of $\pi$ (circles) and $e$ (crosses), i.i.d. elements $w_i$ of $\w$ from $N(0,1)$ (squares) and uniform distribution over $[-1,+1]$ (diamonds). In all cases, the motif richness (measured by relative area covered by Fourier coefficients) monotonically increases with $\nu$ up to $\nu=0.99$ (dashed red vertical line), where there is a phase transition marking the onset of a rapid decline in motif richness. No such behavior can be observed for random $\W$ ($W_{i,j}$ generated i.i.d. from $N(0,1)$ (dashed green lines), where the motif richness is consistently low.

\begin{figure}
\centering
\includegraphics[width=10cm,height=7cm]{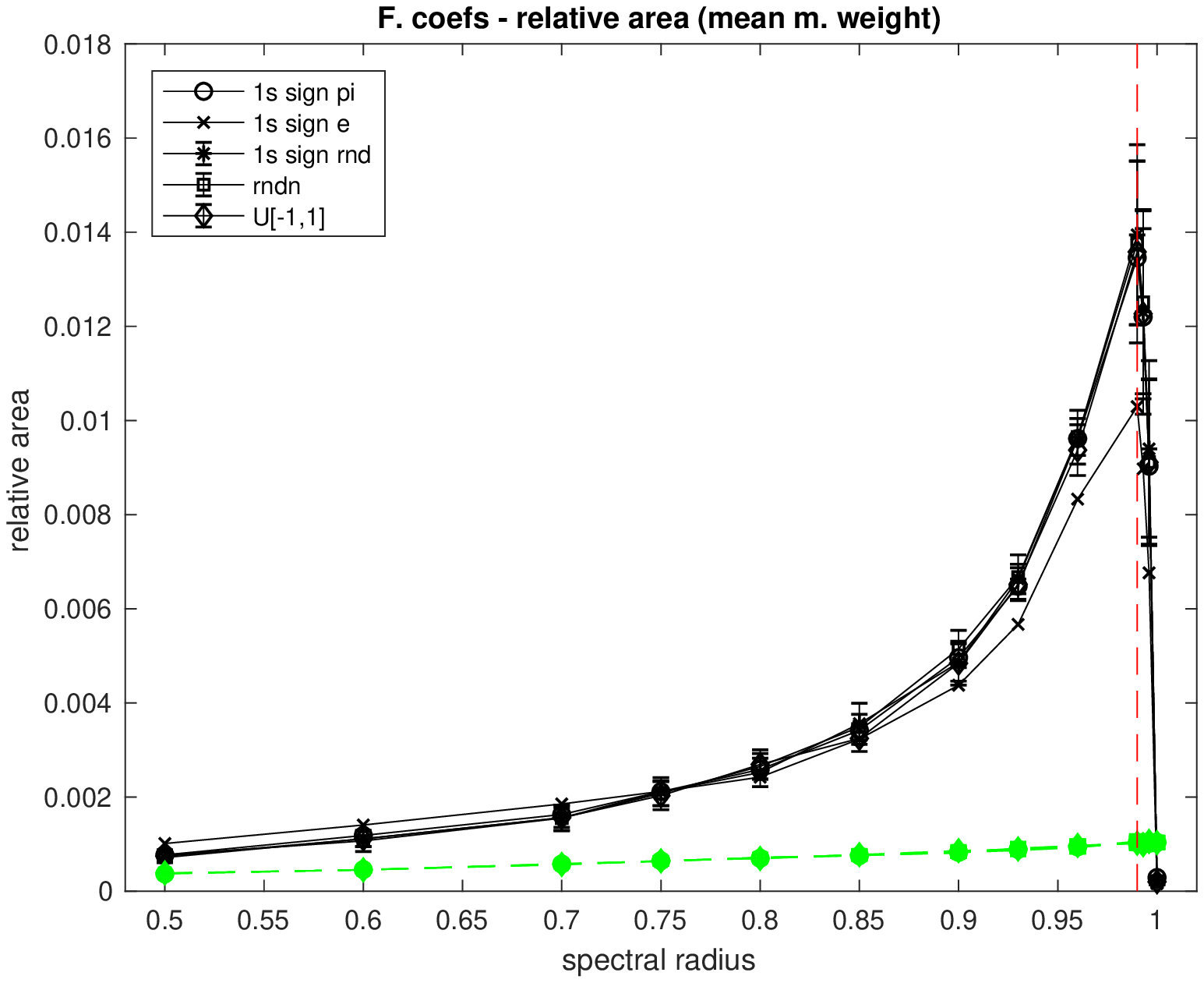}
\caption{Relative area measured using local mean motif weight of Fourier coefficients of temporal kernel motifs. All other settings are equal to those of figure \ref{fig:F_coef_motifs_nu_mean}.}
\label{fig:F_coef_motifs_nu_mean_weight}
\end{figure}

Our second measure of motif richness takes into account motif weights, instead of simply noting whether a particular small cell in the complex plane of Fourier coefficients was visited or not. To that end the motif weights were first normalized to the total sum 1. In each cell $c$ we calculate the mean $\bar q_c$ of the weights $q_{i,k}$ of coefficients $z_{i,k}$ that landed in that cell. 
The {\em relative weighted area} covered by $F$ is the ratio of the accumulated mean weight in cells visited by the coefficients $z_{i,k}$,
$\sum_c \bar q_c$, to the total number of cells.
Figure \ref{fig:F_coef_motifs_nu_mean_weight} shows that the relative weighted area exhibits the same universal behavior as a function of spectral radius $\nu$ of $\W$ as that followed by the relative area (figure \ref{fig:F_coef_motifs_nu_mean}).

\section{Discussion and Conclusion}
\label{sec:Conclusion}
{Parameterized state space models have been used extensively in machine learning community, e.g. in the form of recurrent neural networks. Since learning of the dynamic part is known to be difficult, the key idea of reservoir computation is to restrict learning only to the static readout part from the state space, while keeping the underlying dynamical system fixed. Furthermore, the readout is usually a simple linear mapping. This is very similar in spirit to the idea of kernel machines: Transform the inputs using a fixed mapping (usually only implicitly defined) into another "rich" feature space and only train a linear model in that space, with the dot product as the canonical tool. The key to understanding the workings of kernel machines is to understand the feature space: How are the data mapped from the original space  into the "richer" feature space? What similarity notions does the inner product in the feature space express in terms of the original data space? This paper is the first study suggesting to formalize and rigorously analyze the connection between fixed dynamics in reservoir computation models and fixed kernel-based transformations to feature spaces in kernel machines. So far, theoretical tools at our disposal that would allow us to make statements regarding appropriateness of different settings of dynamic coupling in reservoir computation models have been rather limited. The new framework introduced in this paper allows us to investigate richness of  internal representations of input time series in terms of dynamic states and how they operate to produce desired outputs in terms of matching with a set of temporal motifs defined by the structure of the dynamic coupling.}
Our investigations lead to several rather surprising results:

\begin{enumerate}
\item
The usual strategy of random generation (i.i.d.) of input and dynamic coupling weights in the reservoir of ESN leads to shallow memory time series representations, corresponding to cross-correlation operator with fast exponentially decaying coefficients. This finding is quite robust with respect to distributions with which the ESN parameters are generated. 

\item
Imposing symmetry on coupling weights of the dynamical system yields a constrained dynamic kernel that is a linear combination of trivial kernels matching the input time series with either a straightforward exponentially decaying motif or an exponentially decaying motif of the highest frequency.  

\item
The simple cycle reservoir topology has been empirically demonstrated to have the potential to equal the performance of more complex reservoir settings \cite{Rodan10}. The dynamical system can have high state space dimensionality, but it is specified only through two free parameters, namely a constant coupling weight $r>0$ between consecutive reservoir units in the cycle topology\footnote{
Note that this also automatically makes the spectral radius $\nu$ of $\W$ equal to $r$, so no separate tuning of $\nu$ is needed.} and a constant weight $v>0$ of the input-to-state coupling. The crucial constraint is that the input coupling vector, while all its elements have the same absolute value  $v$, has a-periodically distributed signs. In this paper we have provided rigorous arguments for the need of aperiodic sign distribution in the input coupling, showing that compared with periodic sign patterns, the feature representations of time series in case of a-periodically distributed signs are much richer. In addition, even though such settings of the dynamical system are extremely simple (two free parameters) and completely deterministic\footnote{
The sign distribution can follow binary expansion of an irrational number, such as $\pi$ or $e$.}, 
the number and variety of dynamic motifs of the associated dynamic kernel are far superior to the standard configurations of ESN that rely on stochastic generation of coupling weights.    

\item
By quantifying motif richness of the dynamic kernel associated with cycle reservoir topology, we showed that there is a phase transition in motif richness at spectral radius values close to, but strictly less than  1. This confirms previous findings in the ESN literature on the importance of tuning the dynamical system at the edge of stability \cite{Bertschinger04}.

\end{enumerate}

The arguments in this paper were developed under the assumption of linear dynamical system with linear readout map. However, it has been proved that even linear dynamical systems can be universal, provided they are equipped with polynomial readout maps \cite{Grigoryeva2018,GRIGORYEVA2018_NN,Gonon19}.
In our setting, this corresponds to considering instead of the linear kernel (eq. (\ref{eq:K}))
a polynomial kernel (of some degree $d$),
\begin{equation}
K(\u,\v) 
=
 (\lla \phi(\u),\phi(\v) \rra + a)^d.
\label{eq:K_poly}
\end{equation}
Clearly, memory characteristics of such a kernel will not change with offset $a \in \RR$ or increasing polynomial degree $d$.
By eqs. (\ref{eq:K_m_t}-\ref{eq:tilde_u}), the polynomial kernel can be written as
\begin{equation}
K(\u,\v) 
=
 (\lla \til \u,\til \v \rra + a)^d,
\label{eq:K_poly1}
\end{equation}
where the elements $\til u_i, \til v_i$, $i=1,2,...,N_m$, of  $\til \u, \til \v \in \RR^{N_m}$ are projections of
$\u,\v  \in \RR^\tau$ onto motifs $\m_i \in \RR^\tau$, scaled by the motif weight. Non-linear manipulation of $\til u_i, \til v_i$ can increase the capacity of the readout mapping {\em but only at the level of memory and feature set defined by the motifs}. 
Randomly generated or symmetric reservoir couplings will still lead to constrained shallow memory kernels. We have shown that simple cycle reservoirs tuned at the edge of stability, with aperiodic sign patterns in input coupling are  among the ESN architectures capable of approximating deep memory processes when linear dynamical system and polynomial readout are used.
Of course, when non-linearity is allowed in the dynamical system (for example, by employing a logistic sigmoid transfer function), even randomly generated reservoirs {may} be able to capture deeper memory. 

Our study contributes to the debate about what characteristics of the dynamical system are desirable to make it a `universal' temporal filter capable of producing rich representations of input time series in its state space. Such representations can then be further utilized by readouts, purpose-build for a variety tasks.
\cite{Ozturk2007} hypothesized that 
the distribution of reservoir
activations should have high entropy and suggested that it was desirable for 
the reservoir weight matrix 
to have eigenvalues uniformly distributed inside the
unit circle. In this way the system dynamics would include
uniform coverage of time constants (related to the uniform distribution of the poles)
(Ozturk et al., 2007). Our work suggests a counterargument when linear reservoirs and non-linear readouts are used: Uniform distribution of eigenvalues
inside  the unit circle can be achieved by random generation of the reservoir matrix. However, this leads to a highly constrained set of shallow memory motifs of the associated dynamic kernel that describes how features of time series seen in the past contribute to the production of the model output. On the other hand, a very simple setting of high dimensional dynamical system governed by just two free parameters can achieve a much richer and deeper memory motifs of the dynamic kernel. Note that in this case, the eigenvalues of the reservoir coupling matrix are distributed uniformly along a circle with radius equal to the spectral radius of the reservoir matrix.





\vskip 0.5cm
\noindent
{\bf Acknowledgement:}\\
This work was supported by the European Commission Horizon 2020 Innovative Training Network SUNDIAL (SUrvey Network for
Deep Imaging Analysis and Learning), Project ID: 721463.

\bibliographystyle{plain}
\bibliography{refs}

\appendix

{
\section{Markovian motifs resulting from random dynamical coupling $\W$}
\label{sec:appendix}

In this appendix we demonstrate that the approximations in section \ref{sec:rand_W} of motifs and their weights in the case of random dynamic coupling $\W$ hold for diverse distributions used to generate elements of $\W$.
In particular, in figure 
\ref{fig:motifs_w_1s_rs}
we show
kernel motifs 
obtained from $\W \in \RR^{100 \times 100}$ generated element-wise i.i.d. from $N(0,1)$ and renormalized to largest singular value $\nu=0.995$ (the setting used in section \ref{sec:rand_W}) and
input coupling vector $\w$ generated as a vector of all 1s with randomly flipped signs (in each dimension with equal probability 0.5), renormalized to unit vector. The associated
squared motif weights are presented in figure \ref{fig:eigenvalues_w_1s_rs}.
We also show in figures
\ref{fig:motifs_W_1s_rs_w_1s_rs} and \ref{fig:eigenvalues_W_1s_rs_w_1s_rs} the kernel motifs and eigenvalues of $\Q$ when both $\til \W$ and $\w$ consist of all 1s with signs flipped independently element-wise with probability 0.5 (dynamical coupling renormalized to largest singular value $\nu=0.995$ and $\w$ to unit vector).
In both cases, the Markovian motifs and their weights are almost indistinguishable from those shown in section \ref{sec:rand_W} (figures \ref{fig:motifs_w_randn}
and \ref{fig:eigenvalues_w_randn}).

\begin{figure}
\centering
\includegraphics[width=8cm,height=7.5cm]{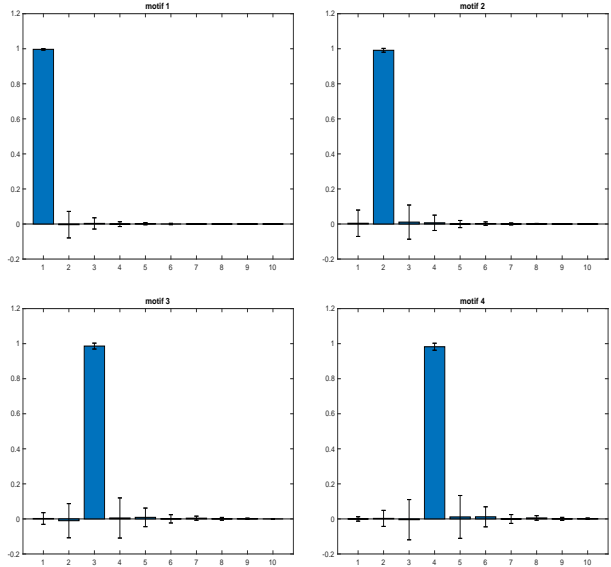}
\caption{The first 10 elements  of the four most dominant kernel motifs corresponding to $\W \in \RR^{100 \times 100}$ generated element-wise i.i.d. from $N(0,1)$ and renormalized to largest singular value $\nu=0.995$. The input coupling $\w$ was
generated as a vector of all 1s with randomly flipped signs (in each dimension with equal probability 0.5).
Shown are the means and standard deviations across 100 joint realizations of $\W$ and $\w$.  
}
\label{fig:motifs_w_1s_rs}
\end{figure}

\begin{figure}
\centering
\includegraphics[width=5.5cm,height=5cm]{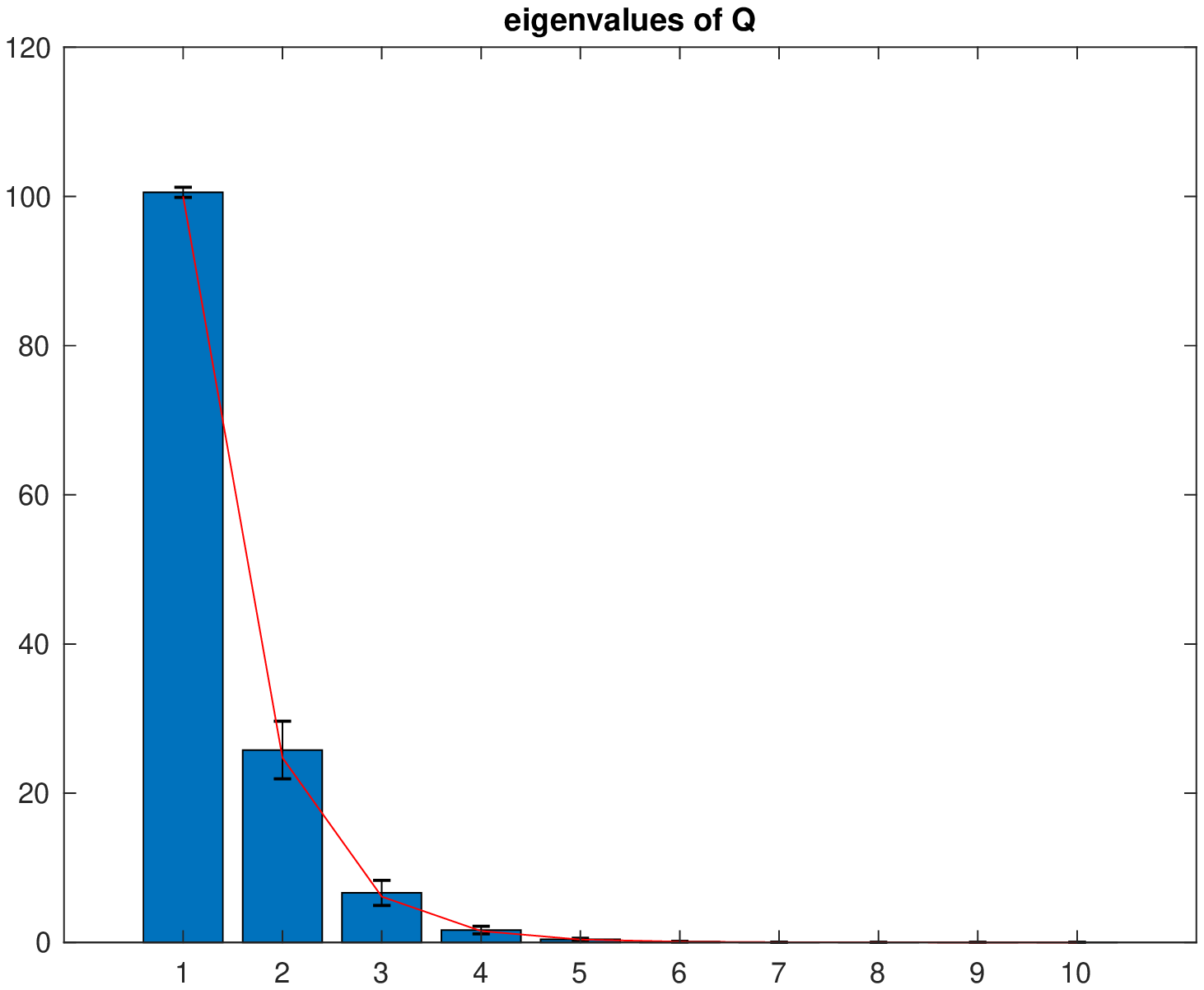}
\caption{Eigenvalues (squared motif weights) of the metric tensor $\Q$ for random setting of the dynamical system (\ref{eq:state}) as described in fig. \ref{fig:motifs_w_1s_rs}. Solid bars correspond to the means of the actual eigenvalues $\lambda_i$ across the 100 realizations of $\W$ and $\w$ (also shown are standard deviations). The theoretically predicted values (eq. (\ref{eq:hat_lambda_i})) are shown as the red line. 
}
\label{fig:eigenvalues_w_1s_rs}
\end{figure}

\begin{figure}
\centering
\includegraphics[width=8cm,height=7.5cm]{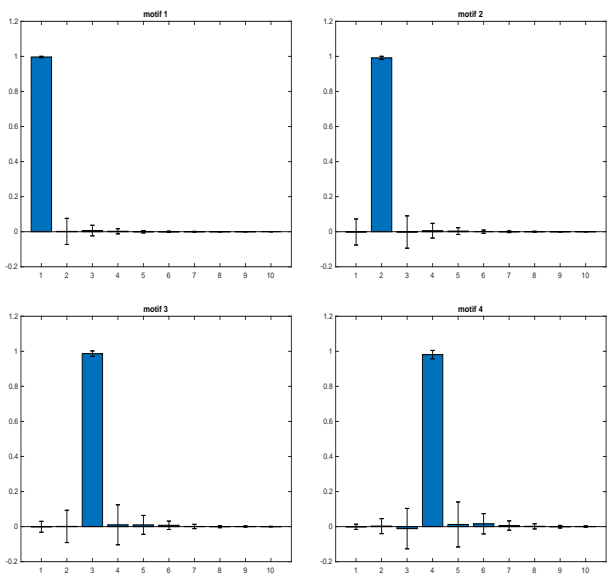}
\caption{The first 10 elements  of the four most dominant kernel motifs corresponding to $\W \in \RR^{100 \times 100}$ and $\w$, 
both consisting of all 1s with signs flipped independently element-wise with probability 0.5.
$\W$ was renormalized to largest singular value $\nu=0.995$. 
Shown are the means and standard deviations across 100 joint realizations of $\W$ and $\w$.  
}
\label{fig:motifs_W_1s_rs_w_1s_rs}
\end{figure}

\begin{figure}
\centering
\includegraphics[width=5.5cm,height=5cm]{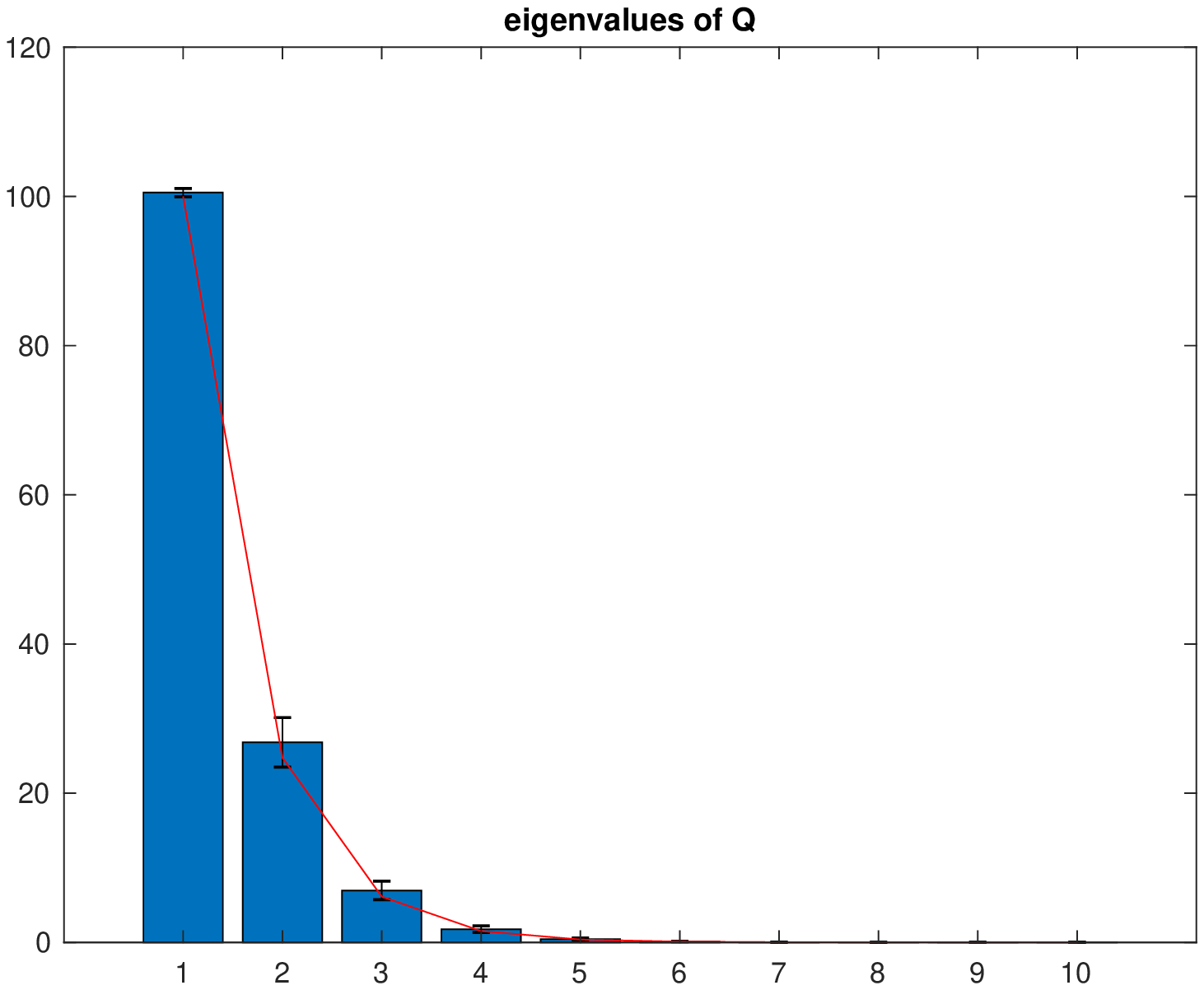}
\caption{Eigenvalues (squared motif weights) of the metric tensor $\Q$ for random setting of the dynamical system (\ref{eq:state}) as described in fig. \ref{fig:motifs_W_1s_rs_w_1s_rs}. Solid bars correspond to the means of the actual eigenvalues $\lambda_i$ across the 100 realizations of $\W$ and $\w$ (also shown are standard deviations). The theoretically predicted values (eq. (\ref{eq:hat_lambda_i})) are shown as the red line. 
}
\label{fig:eigenvalues_W_1s_rs_w_1s_rs}
\end{figure}

}

\end{document}